%% file: TPAMI_2022.tex
\documentclass[10pt,journal,compsoc]{IEEEtran}
\ifCLASSOPTIONcompsoc
  \usepackage[nocompress]{cite}
\else
  \usepackage{cite}
\fi
\ifCLASSINFOpdf
\else
\fi
\usepackage[numbers]{natbib}
\usepackage[utf8]{inputenc} % allow utf-8 input
\usepackage[T1]{fontenc}    % use 8-bit T1 fonts
\usepackage{url}            % simple URL typesetting
\usepackage{booktabs}       % professional-quality tables
\usepackage{amsfonts}       % blackboard math symbols
\usepackage{amsmath}
\usepackage{nicefrac} 
\usepackage{amssymb}% compact symbols for 1/2, etc.
\usepackage{microtype}      % microtypography
\usepackage{xcolor}         % colors
\usepackage{graphicx}
\usepackage{adjustbox}
\usepackage{multirow}
\usepackage{multicol}
\usepackage{bbm}
\usepackage{mathtools}
\usepackage{amsthm}
\usepackage{array,booktabs}
\usepackage{enumitem}
\usepackage{wrapfig}
\usepackage{bm}
\usepackage{listings}
\usepackage{color}
\usepackage{pifont}% http://ctan.org/pkg/pifont
\usepackage{multirow}
\usepackage{rotating}
\usepackage{hyperref}
\newcommand{\cmark}{\ding{51}}%
\newcommand{\xmark}{\ding{55}}%
\newtheorem{theorem}{Theorem}
\numberwithin{theorem}{section} 
\newtheorem{definition}{Definition}

\numberwithin{lemma}{section}

\begin{document}
\title{ImDrug: A Benchmark for Deep Imbalanced Learning in AI-aided Drug Discovery}

\author{Lanqing Li,~\IEEEmembership{Student Member,~IEEE},
        Liang Zeng,
        Ziqi Gao,
        Shen Yuan,
        Yatao Bian,
        Bingzhe Wu,
        Hengtong Zhang,
        Yang Yu,
        Chan Lu,
        Zhipeng Zhou,~\IEEEmembership{Student Member,~IEEE},
        Hongteng Xu,~\IEEEmembership{Member,~IEEE},
        Jia Li,
        Peilin Zhao,
        and Pheng-Ann Heng,~\IEEEmembership{Senior Member,~IEEE}% <-this % stops a space
\IEEEcompsocitemizethanks{
\IEEEcompsocthanksitem The first four authors contributed equally to this work. Lanqing Li is the corresponding author. E-mail: lanqingli1993@gmail.com 
\IEEEcompsocthanksitem Lanqing Li, Yatao Bian, Bingzhe Wu, Hengtong Zhang, Yang Yu, Chan Lu, Peilin Zhao are with Tencent AI Lab.
% yatao.bian@gmail.com, wubingzhe@pku.edu.cn, htzhang.work@gmail.com, kevinyyu@tencent.com, chanlu009@gmail.com, masonzhao@tencent.com.\protect\\
% E-mail: see http://www.michaelshell.org/contact.html
\IEEEcompsocthanksitem Lanqing Li and Pheng-Ann Heng are with Department of Computer Science and Engineering, The Chinese University of Hong Kong.
% E-mail: \{lqli, pheng\}@cse.cuhk.edu.hk.\protect\\
\IEEEcompsocthanksitem Liang Zeng is with Institute for Interdisciplinary Information Sciences (IIIS), Tsinghua University.
% E-mail: zengl18@mails.tsinghua.edu.cn.\protect\\
\IEEEcompsocthanksitem Jia Li and Ziqi Gao are with Hong Kong University of Science and Technology (Guangzhou) and Hong Kong University of Science and Technology.
% E-mail: zgaoat@connect.ust.hk and jialee@ust.hk.\protect\\
\IEEEcompsocthanksitem Shen Yuan and Hongteng Xu are with Gaoling School of Artificial Intelligence, Renmin University of China. 
% E-mail: \{shenyuan721, hongtengxu\}@ruc.edu.cn.\protect\\
\IEEEcompsocthanksitem Zhipeng Zhou is with 
% School of Computer Science and Technology, 
University of Science and Technology of China. % E-mail: zzp1994@mail.ustc.edu.cn.
}% <-this % stops an unwanted space
}
\IEEEtitleabstractindextext{%
\input{Abstract}
% \begin{abstract}
% The abstract goes here.
% \end{abstract}

}
\maketitle
\input{Sections-TPAMI/1-Introduction}

\input{Sections-TPAMI/2-Related_work}
\input{Sections-TPAMI/3-Overview}
\input{Sections-TPAMI/4-Experiments}
\input{Sections-TPAMI/5-Conclusion}

\ifCLASSOPTIONcompsoc
  \section*{Acknowledgments}
\else
  \section*{Acknowledgment}
\fi

This work was partially supported by Tencent AI Lab Rhino-Bird Focused Research Program JR202103.

\ifCLASSOPTIONcaptionsoff
  \newpage
\fi



\bibliographystyle{IEEEtranN}
\bibliography{TPAMI_2022}


% \end{thebibliography}


\begin{IEEEbiography}[\adjustbox{width=1.1in,height=1.25in,trim={.05\width} {.\height} {0.05\width} {0.12\height},clip,keepaspectratio}{\includegraphics{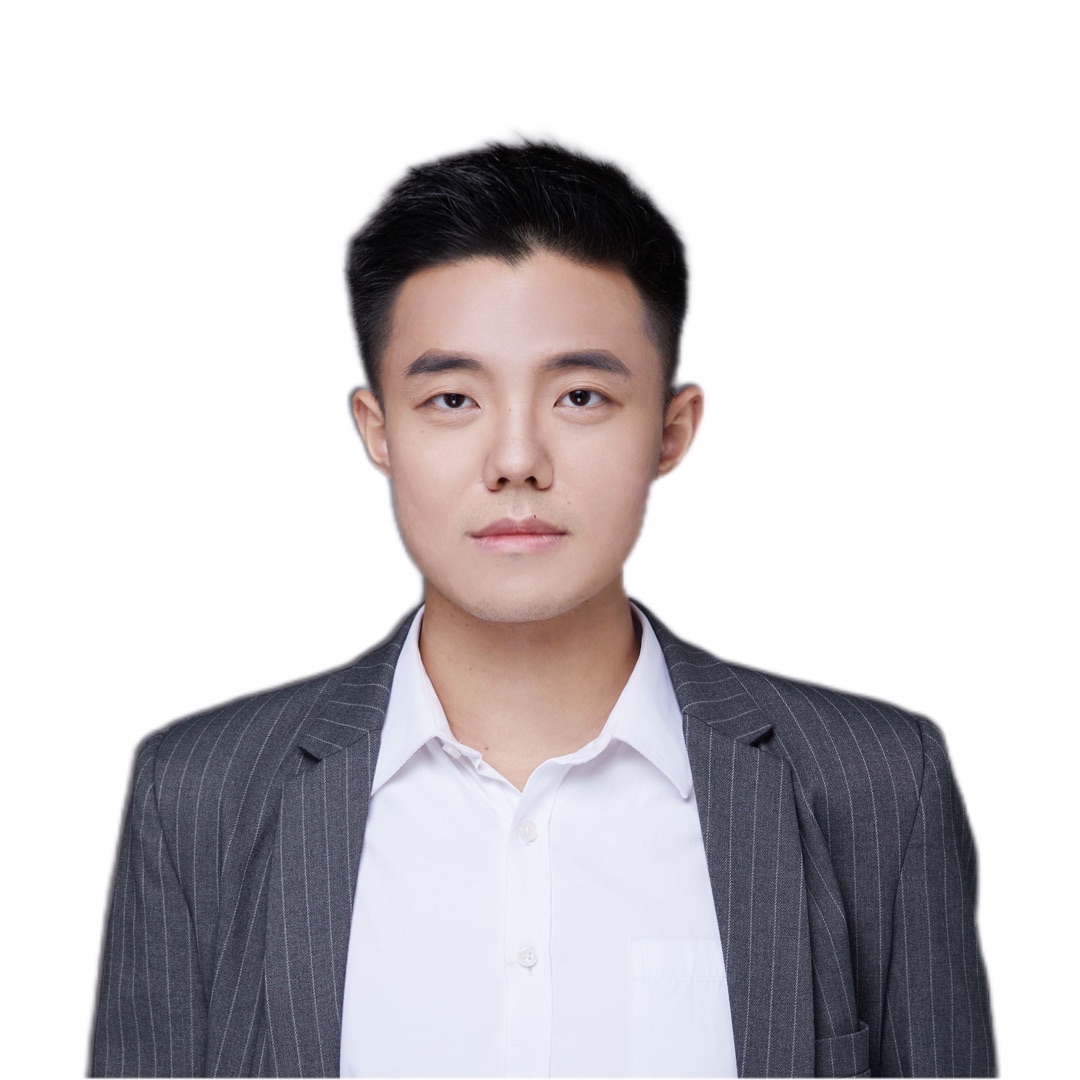}}]{Lanqing Li} is currently a senior research scientist at Tencent AI Lab, working on machine learning and its applications in drug discovery and autonomous control. He is also working towards the Ph.D. degree at the department of computer science and engineering, The Chinese University of Hong Kong. He holds a Master's Degree in Physics from The University of Chicago in 2017 and received his Bechelor's Degree in Physics from MIT in 2015. His research interests span a broad spectrum of machine/deep learning, AI-aided drug discovery and reinforcement learning, with a keen focus on algorithmic generalization and robustness. He has published papers in top venues, including ICLR, ICML, NeurIPS, AAAI, SIGIR, and Physics Review, and was a gold medalist with Prize of Best Score in Theory at the 42nd International Physics Olympiad in 2011. He has been invited as a PC member or reviewer for top international conferences and journals,including NeurIPS, ICML, AAAI and IJCAI.
\end{IEEEbiography}

\vspace*{-11mm}

\begin{IEEEbiography}
[{\includegraphics[width=1in,height=1.25in,clip,keepaspectratio]{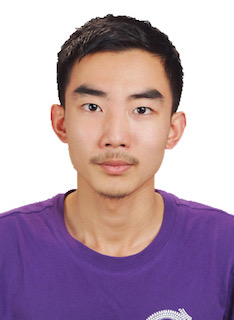}}]{Liang Zeng} is currently a fifth-year PhD student in the Institute for Interdisciplinary Information Sciences(IIIS), Tsinghua University.  Before that he received B.E. degree from the University of Electronic Science and Technology of China. His current research interest focuses on Graph Neural Networks and Time-series Data.
\end{IEEEbiography}

\vspace*{-9mm}

\begin{IEEEbiography}
[{\includegraphics[width=1in,height=1.25in,clip,keepaspectratio]{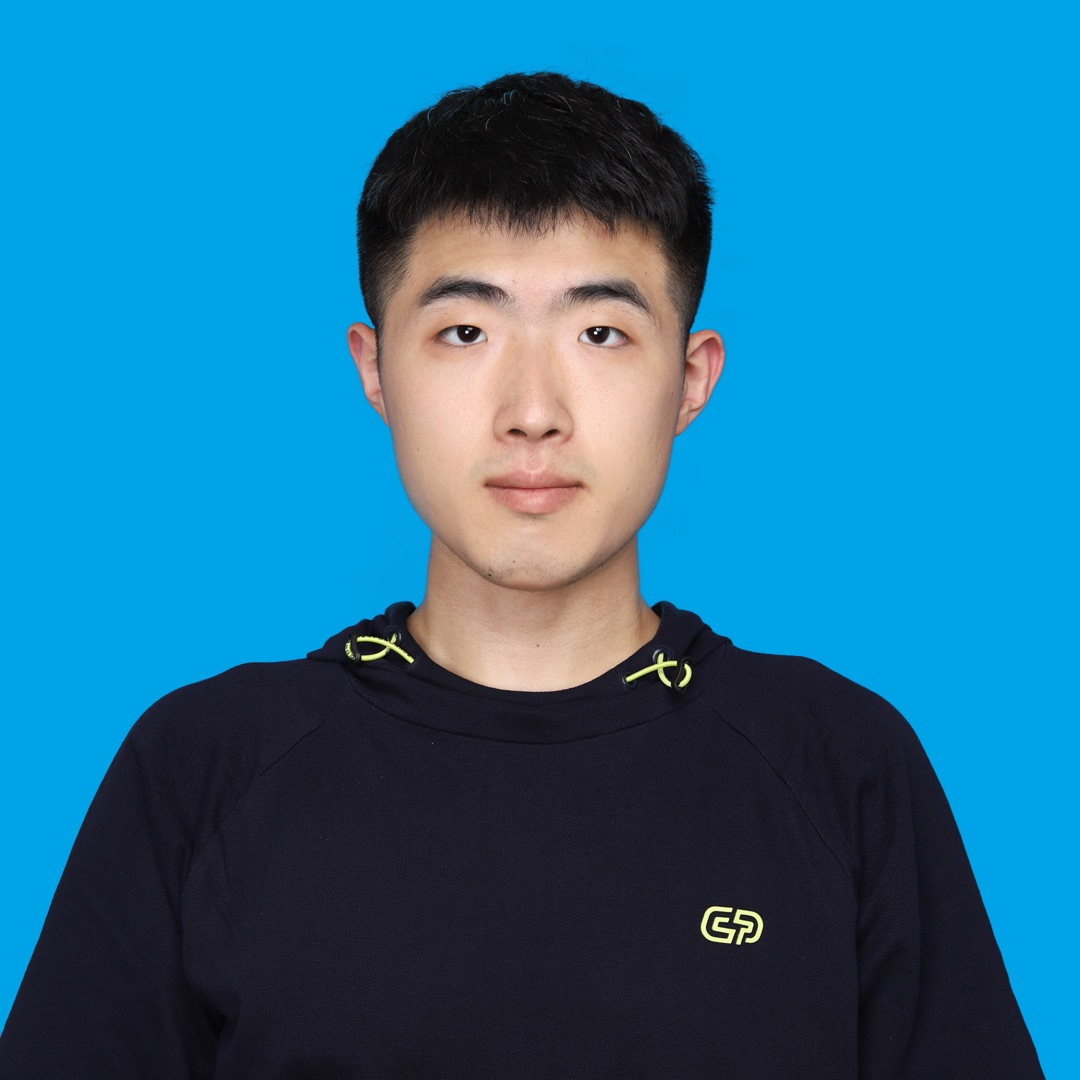}}]{Ziqi Gao} is currently pursuing a Ph.D. Degree at Hong Kong University of Science and Technology. He received his B.S. and M.S. from Huazhong University of Science \& Technology and Tsinghua University, respectively. His research interests focus on machine learning and deep graph learning.
\end{IEEEbiography}

\vspace*{-11mm}

\begin{IEEEbiography}
[{\includegraphics[width=1in,height=1.25in,clip,keepaspectratio]{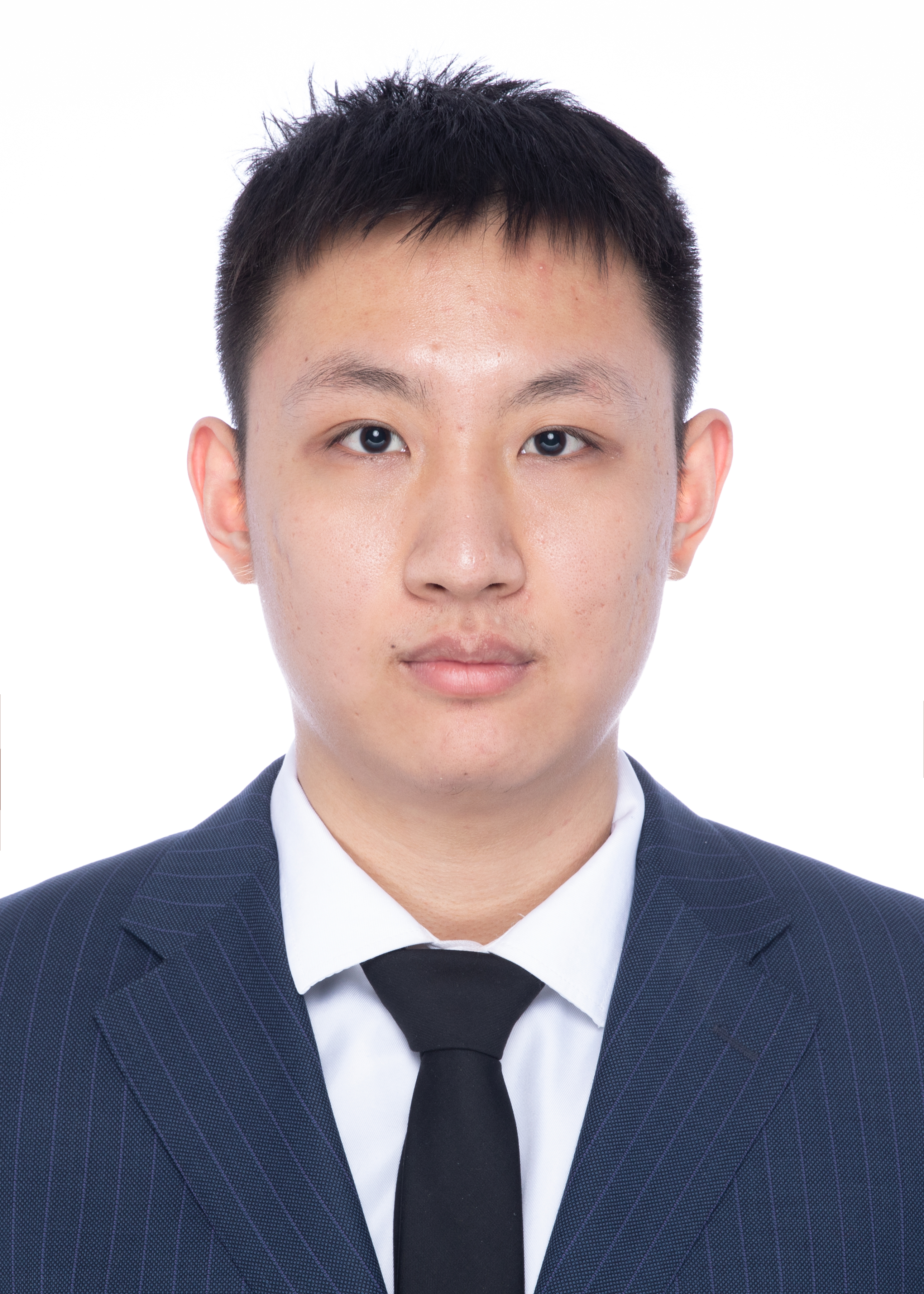}}]{Shen Yuan} is a Ph.D student at Gaoling School of Artificial Intelligence, Renmin University of China, under the supervision of Prof. Hongteng Xu. He majors in Artificial Intelligence. His research interests lie in the application of machine learning, specifically in the area of drug retrosynthesis. He currently focus on the single-retrosynthesis prediction task.
\end{IEEEbiography}

\vspace*{-9mm}

\begin{IEEEbiography}
[{\includegraphics[width=1in,height=1.25in,clip,keepaspectratio]{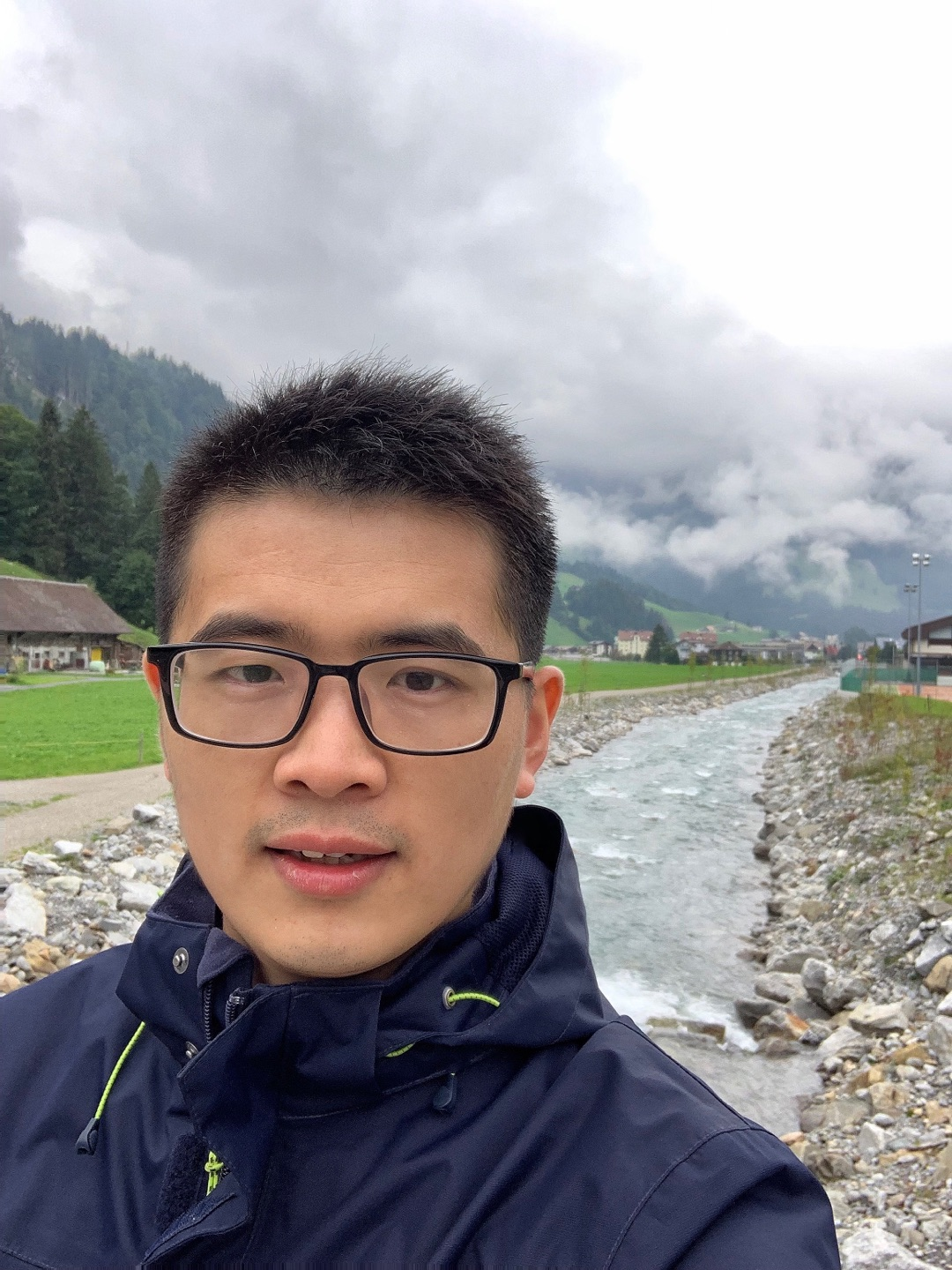}}]{Yatao Bian} is a senior researcher of Machine Learning Center in Tencent AI Lab.  He received the Ph.D. degree from the Institute for Machine Learning at ETH Zurich. He has been an associated Fellow of the Max Planck ETH Center for Learning Systems since June 2015. Before the Ph.D. program he obtained both of his M.Sc.Eng. and B.Sc.Eng. degrees from Shanghai Jiao Tong University. He is now working on graph representation learning, interpretable ML, and drug AI. He has won the National Champion in AMD China Accelerated Computing Contest 2011-2012. He has published several papers on machine learning top conferences/Journals such as NeurIPS, ICML, ICLR, T-PAMI, AISTATS etc. He has served as a reviewer/PC for conferences like ICML, NeurIPS, AISTATS, CVPR, AAAI, STOC and journals such as JMLR and T-PAMI.
\end{IEEEbiography}

\vspace*{-8mm}

\begin{IEEEbiography}
[{\includegraphics[width=1in,height=1.25in,clip,keepaspectratio]{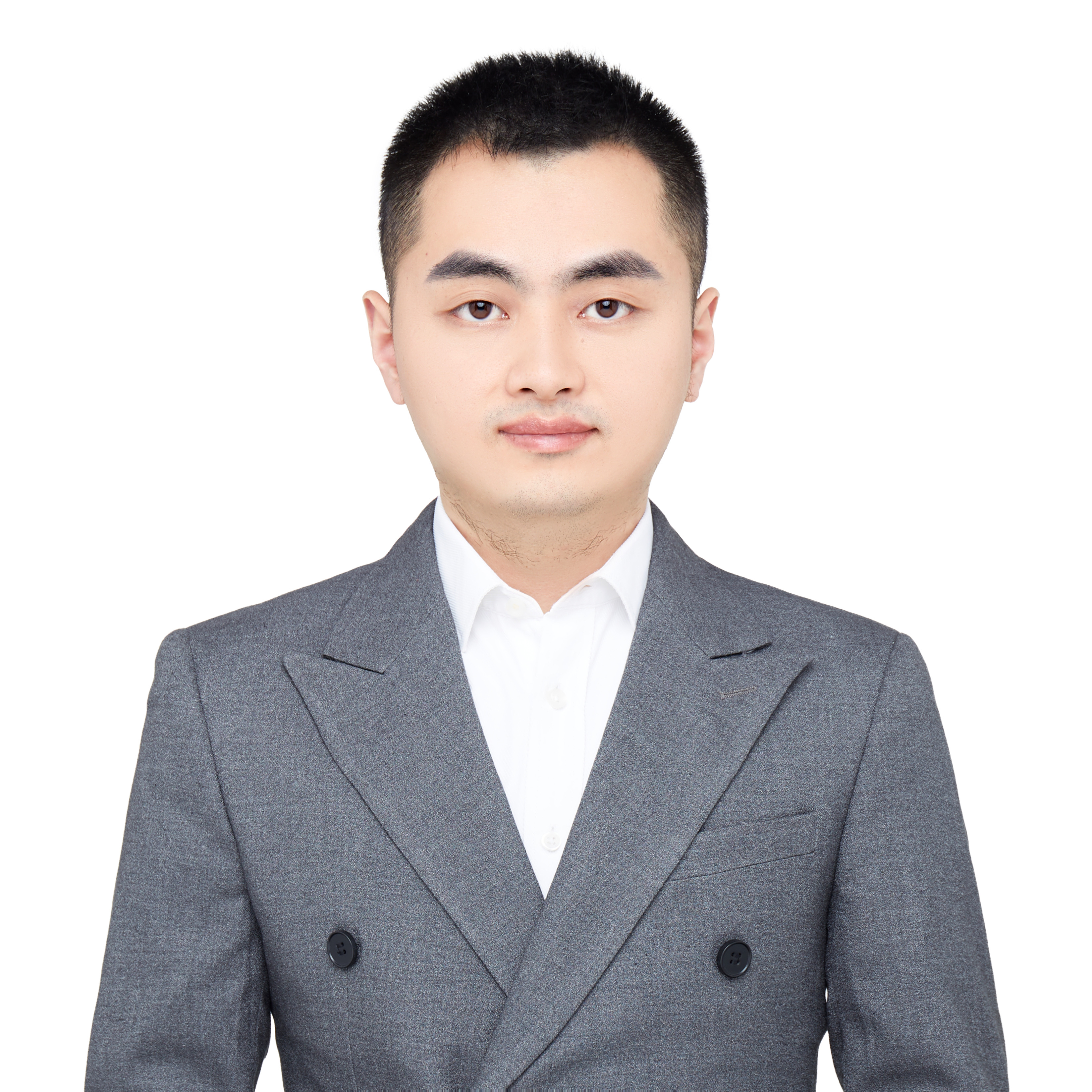}}]{Bingzhe Wu} is currently a senior researcher at Tencent AI LAB. His research interests Iie in various aspects of Trustworthy AI including privacy-preserving machine learning and . He graduated from the Department of Mathematics of Peking University with a bachelor's degree, and received a Ph.D. from the Department of Computer Science of Peking University. He has published more than twenty papers in top conferences such as NeurlPS, HPCA, ISCA, CVPR, etc. He has won many honorary titles such as ACM China Information Security Society Outstanding Doctor, China Electronic Education Association Outstanding Doctor, Apple Doctoral Scholarship (only one in mainland China every year), and Beijing Outstanding Graduate. He also participated in the development of the IEEE P2830 shared machine learning standard as committee secretary.
\end{IEEEbiography}

\vspace*{-8mm}

\begin{IEEEbiography}
[{\includegraphics[width=1in,height=1.25in,clip,keepaspectratio]{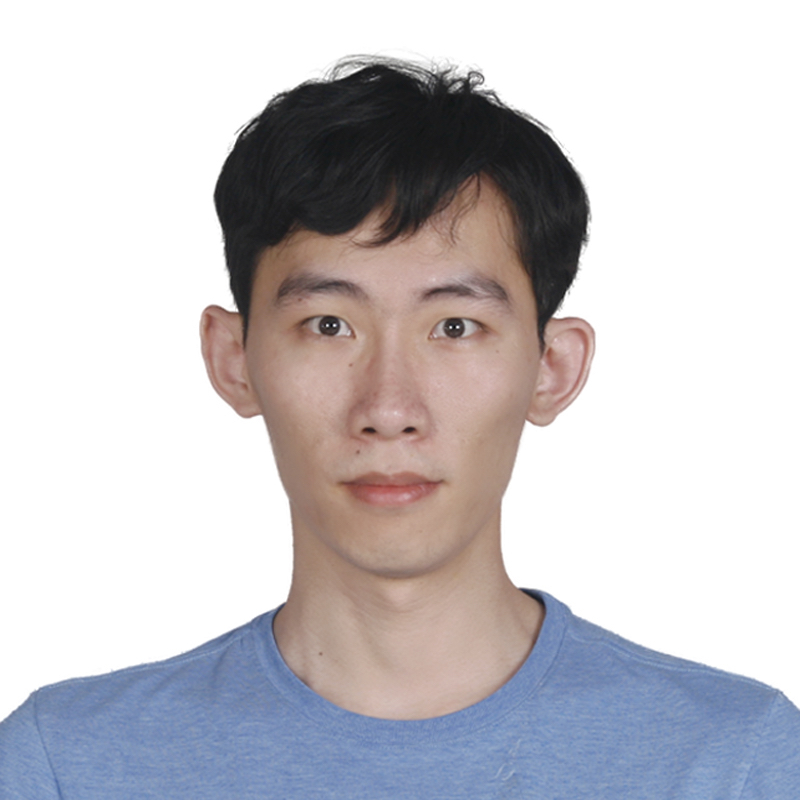}}]{Hengtong Zhang} is a senior researcher at Tencent AI lab. He received his Ph.D. degree from the Department of Computer Science and Engineering at SUNY at Buffalo. Before that he received B.E. degree from the School of Computer at Beijing University of Posts and Telecommunications. He is broadly interested in machine learning, data mining and  security with a focus on adversarial learning, data trustworthy analysis, text mining, and reinforcement learning.
\end{IEEEbiography}

\newpage

\begin{IEEEbiographynophoto}
{Yang Yu} obtained his Ph.D. degree from the University of Hong Kong in 2015.  His research in Hong Kong was focused on the total synthesis of natural products with anti-malaria activity. He moved to University of Texas at San Antonio to continue his postdoctoral research with Prof. Michael Doyle in the development of new organic reactions and medicinal chemistry. In early 2018, Yang went back to LKS Faculty of Medicine, HKU, and spent a short period where he designed new PET-CT tracers. He moved to Global Health Drug Discovery Institute in Beijing to work on drug discovery projects in malaria and tuberculosis diseases. Dr. Yu has published several research articles, reviews in high-level international journals. His research interests include applications of AI in drug discovery, chemistry industry and material science.
\end{IEEEbiographynophoto}

\vspace*{-20mm}

\begin{IEEEbiographynophoto}
{Chan Lu} is currently a senior researcher at Tencent AI lab.  She received her M.E. degree from Shanghai University of Finance \& Economics in 2017. Before that she received B.E. degree from Southeast University in 2014. Her current research interests include machine learning, data mining and big data analysis.
\end{IEEEbiographynophoto}

\vspace*{-11mm}

\begin{IEEEbiography}
[{\includegraphics[width=1in,height=1.25in,clip,keepaspectratio]{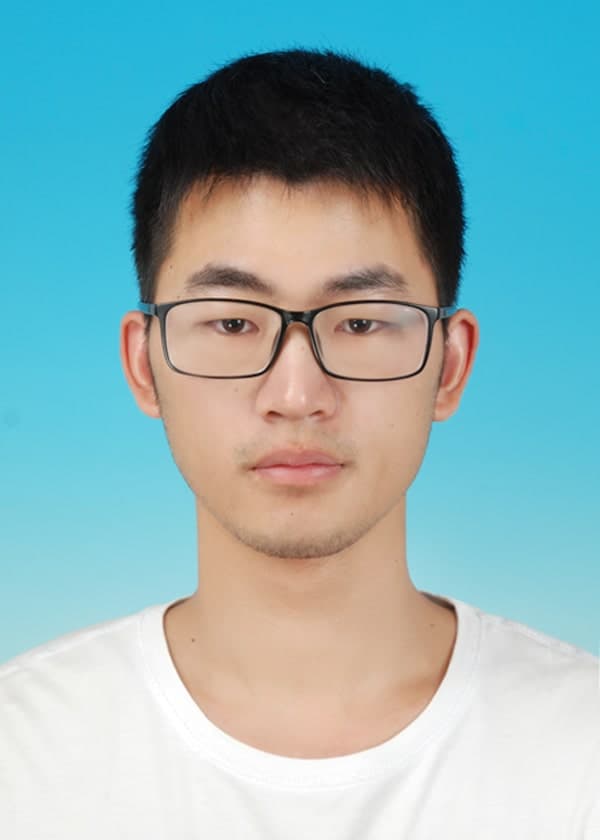}}]{Zhipeng Zhou} (S'21) is currently a PhD Candidate in the School of Computer Science and Technology, University of Science and Technology of China. His research interests include machine learning and its application on wireless sensing. 
\end{IEEEbiography}

\vspace*{-11mm}

\begin{IEEEbiography}
[{\includegraphics[width=1in,height=1.25in,clip,keepaspectratio]{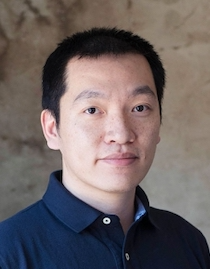}}]{Hongteng Xu} is is an Associate Professor (Tenure-Track) in the Gaoling School of Artificial Intelligence, Renmin University of China. From 2018 to 2020, he was a senior research scientist in Infinia ML Inc. In the same time period, he is a visiting faculty member in the Department of Electrical and Computer Engineering, Duke University. He received his Ph.D. from the School of Electrical and Computer Engineering at Georgia Institute of Technology (Georgia Tech) in 2017. His research interests include machine learning and its applications, especially optimal transport theory, sequential data modeling and analysis, deep learning techniques, and their applications in computer vision and data mining.
\end{IEEEbiography}

\vspace*{-11mm}

\begin{IEEEbiography}
[{\includegraphics[width=1in,height=1.25in,clip,keepaspectratio]{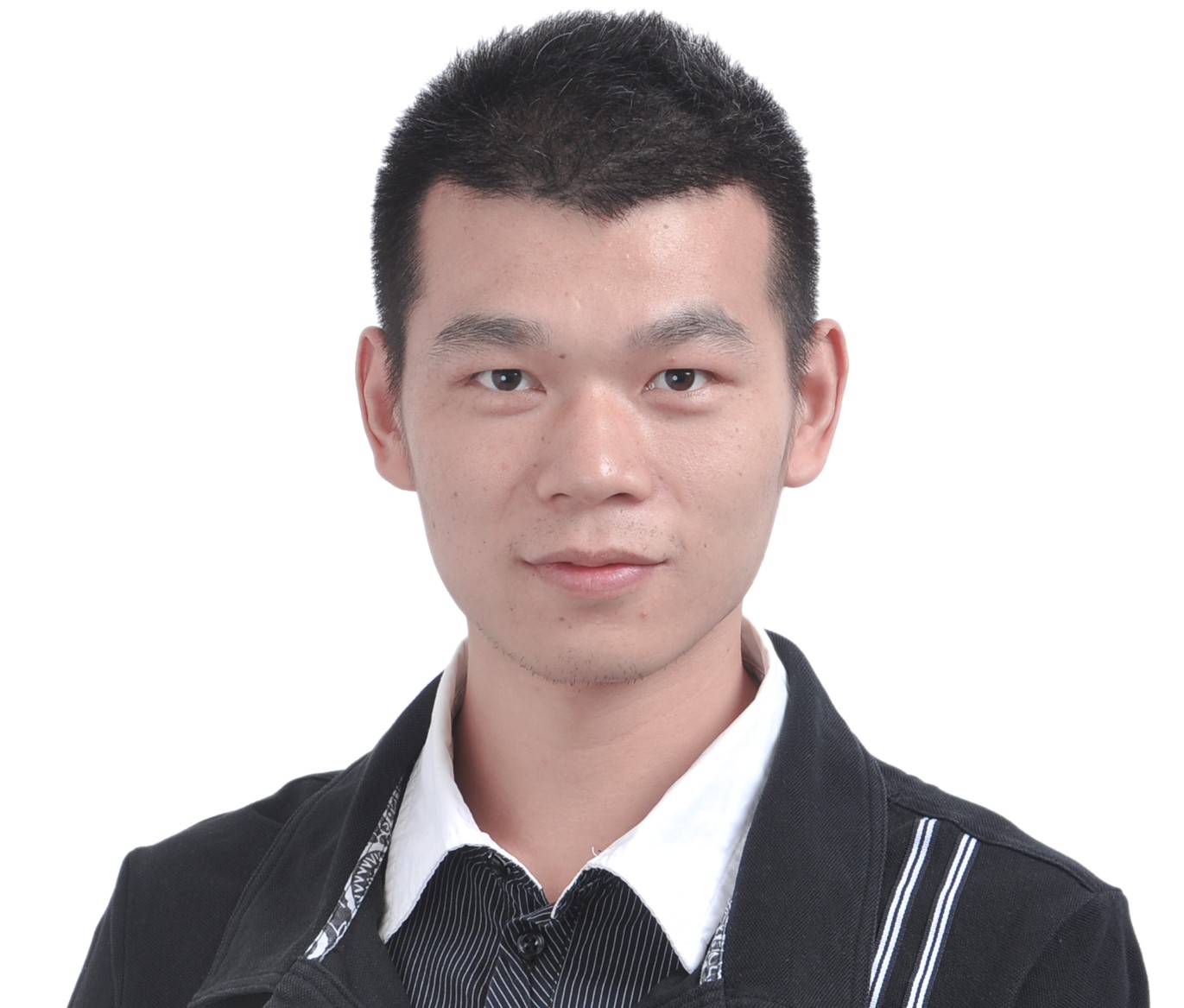}}]{Jia Li} is an assistant professor at Hong Kong University of Science and Technology (Guangzhou) and an affiliated assistant professor at Hong Kong University of Science and Technology. He received the Ph.D. degree at Chinese University of Hong Kong in 2021. His research interests include machine learning, data mining and deep graph learning. He has published several papers as the leading author in top conferences such as KDD, WWW and NeurIPS.
\end{IEEEbiography}

\vspace*{-11mm}

\begin{IEEEbiography}
[{\includegraphics[width=1in,height=1.25in,clip,keepaspectratio]{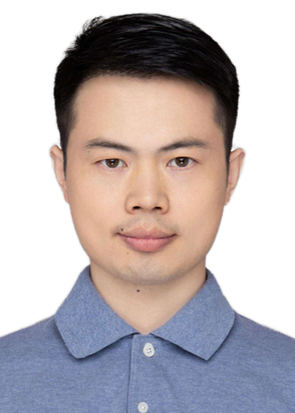}}] {Peilin Zhao} is currently a principal researcher at Tencent AI Lab in China. Previously, he worked at Rutgers University, Institute for Infocomm Research, and Ant Group. His research interests include: online learning, recommendation systems, automatic machine learning, deep graph learning, and reinforcement learning, among others. He has been invited to serve as area chair or associate editor at leading international conferences and journals such as ICML, TPAMI, etc. He received a bachelor's degree in mathematics from Zhejiang University, and a Ph.D. degree in computer science from Nanyang Technological University.
\end{IEEEbiography}

\newpage
\vspace*{-10mm}

\begin{IEEEbiography}
[{\includegraphics[width=1in,height=1.25in,clip,keepaspectratio]{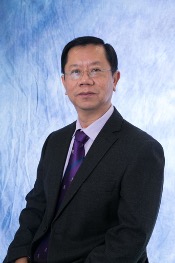}}]{Pheng Ann Heng} is a Professor at the Department of Computer Science and Engineering and Director of Institute of Medical Intelligence and XR at The Chinese University of Hong Kong. His research interests include AI/XR for medical applications, surgical simulation, visualization, computer graphics, and human-computer interaction. He has published over 550 publications in these areas with 33500+ Google citations.
\end{IEEEbiography}


%\newpage


%\vfill


\end{document}

%%%%%%%%%%%%%%%%%%%%%%%

\clearpage
\appendix
\input{Appendix/Appendix-A-Accessibility}
\input{Appendix/Appendix-B-Proof}
\input{Appendix/Appendix-C-Usage}
\input{Appendix/Appendix-D-Dataset}
\input{Appendix/Appendix-E-Baseline}
\input{Appendix/Appendix-F-Backbone}
\input{Appendix/Appendix-G-Implementation}
\input{Appendix/Appendix-I-Metrics}
\input{Appendix/Appendix-J-Train-curve}
% \bibliography{neurips_data_2022}
% \bibliographystyle{unsrtnat}
\bibliographystyle{IEEEtranN}
\bibliography{TPAMI_2022}

%% file: Abstract.tex
\begin{abstract}
    The last decade has witnessed a prosperous development of computational methods and dataset curation for AI-aided drug discovery (AIDD). However, real-world pharmaceutical datasets often exhibit highly imbalanced distribution, which is overlooked by the current literature but may severely compromise the fairness and generalization of machine learning applications. Motivated by this observation, we introduce ImDrug, a comprehensive benchmark with an open-source Python library which consists of 4 imbalance settings, 11 AI-ready datasets, 54 learning tasks and 16 baseline algorithms tailored for imbalanced learning. It provides an accessible and customizable testbed for problems and solutions spanning a broad spectrum of the drug discovery pipeline, such as molecular modeling, drug-target interaction and retrosynthesis. We conduct extensive empirical studies with novel evaluation metrics, to demonstrate that the existing algorithms fall short of solving medicinal and pharmaceutical challenges in the data imbalance scenario. 
    We believe that ImDrug opens up avenues for future research and development, on real-world challenges at the intersection of AIDD and deep imbalanced learning\footnote{Source code and datasets are available at: \url{https://github.com/DrugLT/ImDrug}.}.
\end{abstract}

%% file: Sections-TPAMI/1-Introduction.tex
\section{Introduction}
\label{sec:introduction}
\IEEEPARstart{O}{n} average, it costs over a decade and up to about 3 billion USD to bring a new drug to the market~\cite{pushpakom2019drug}.
% ~\cite{pushpakom2019drug,chan2019advancing}. 
The current drug discovery pipeline devised by domain experts, which largely relies on labor-intensive wet-lab trials, delivers a commercialized drug at only approximately 10\% success rate~\cite{tamimi2009drug} despite substantial investment. AI-aided drug discovery (AIDD) as an emerging area of research, has the potential in revolutionizing the pharmaceutical industry by offering highly efficient and data-driven computational tools to identify new compounds and model complex biochemical mechanisms, thus considerably reducing the cost of drug discovery~\cite{schneider2020rethinking}. A key to the success of this dry-lab paradigm is deep learning. To date, AI driven by deep neural networks has significantly advanced the state-of-the-art in pharmaceutical applications ranging from de novo drug design~\cite{popova2018deep,simonovsky2018graphvae}, ADMET prediction~\cite{wu2018moleculenet,rong2020self}, retrosynthesis~\cite{DBLP:journals/nature/SeglerPW18,coley2018machine,schwaller2021mapping,lu2022unified}, protein folding and design~\cite{jumper2021highly,baek2021accurate,bryant2021deep} to virtual screening~\cite{hu2016large,karimi2019deepaffinity,lim2019predicting}.

\begin{figure}
    \centering
    \includegraphics[width=\columnwidth]{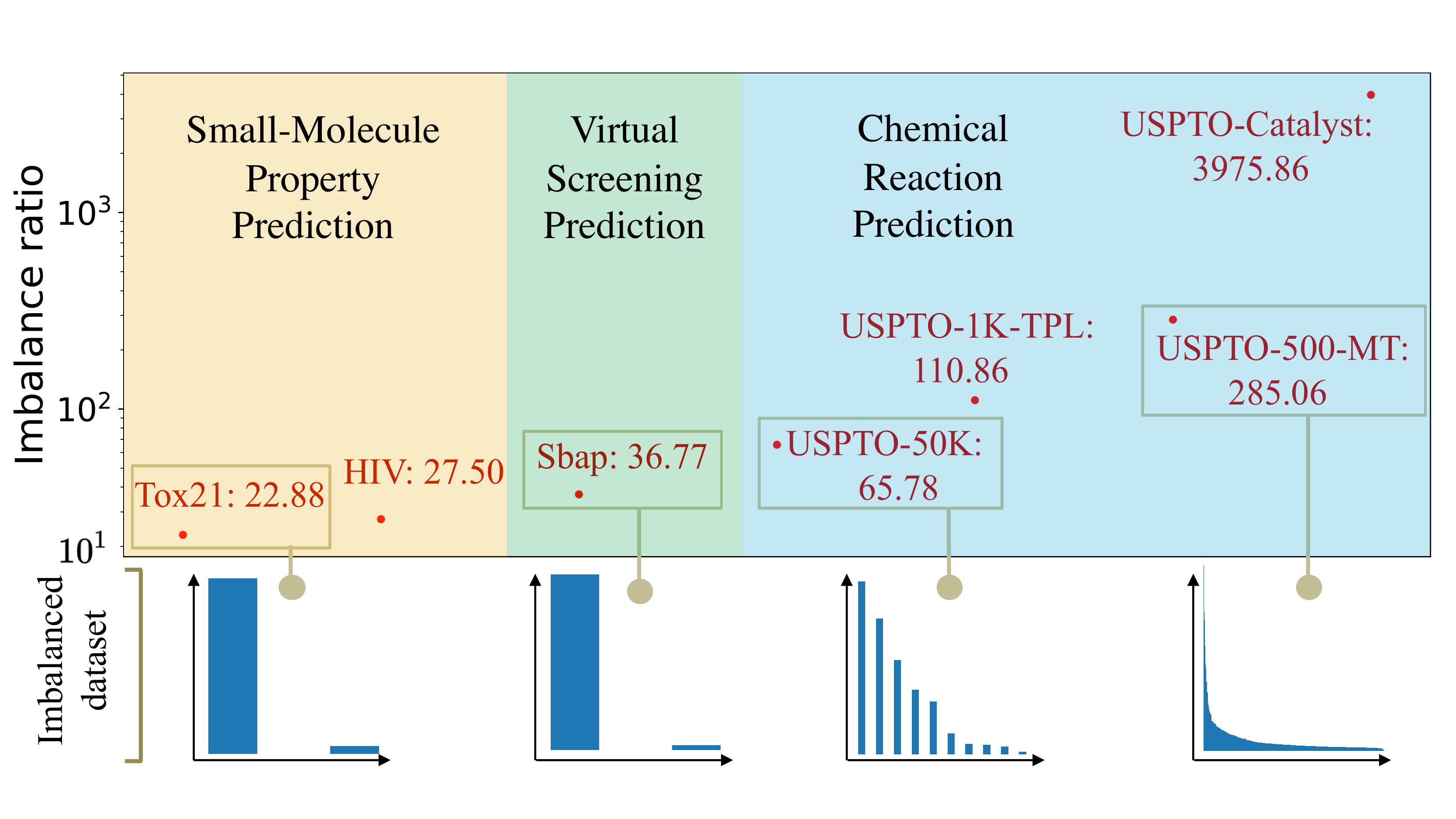}
    \vspace*{-\baselineskip}
    \caption{\textbf{Label distribution and imbalance ratio of the ImDrug datasets and tasks.}}
    \vspace*{-\baselineskip}
    \label{fig:data}
\end{figure}

Despite the success of deep learning in AIDD, by examining a myriad of medicinal chemistry databases and benchmarks~\cite{wu2018moleculenet,schwaller2021mapping,lu2022unified,lowe2012extraction,mendez2019chembl,huang2021therapeutics,2022arXiv220109637J}, we observe that these curated data repositories ubiquitously exhibit imbalanced distributions regardless of the specific tasks and domains, shown in Fig.~\ref{fig:data}. This observation is reminiscent of the power-law scaling in networks~\cite{barabasi1999emergence} and the Pareto principle~\cite{sanders1987pareto}, which poses significant challenges for developing unbiased and generalizable AI algorithms~\cite{DBLP:journals/corr/abs-2110-01167}. 
% (\emph{aka} trustworthy) intro on imbalanced learning

\begin{figure*}[t]
    \centering
    \includegraphics[clip,width=\linewidth]{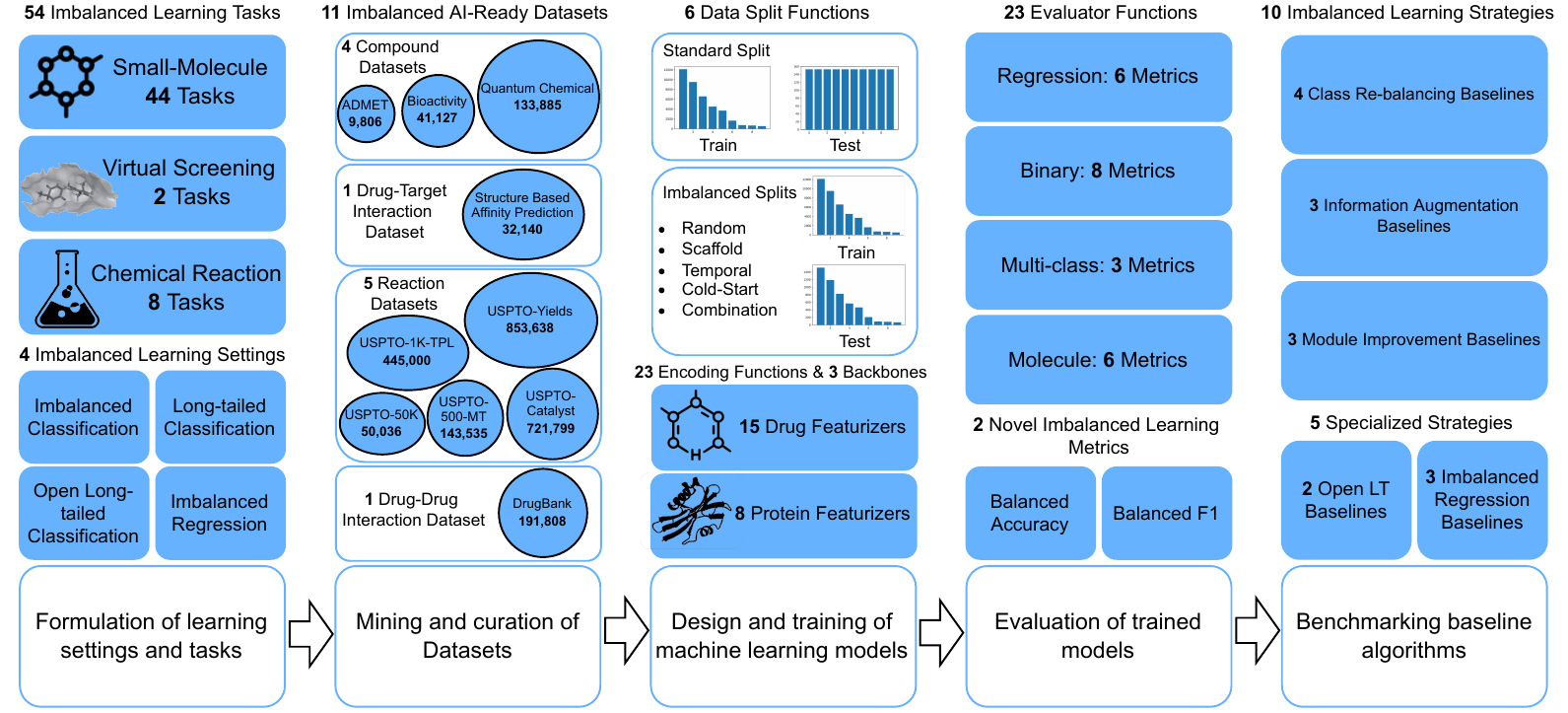}
    \caption{\textbf{Overview of the ImDrug pipeline.} ImDrug provides datasets, modules and utilities for training and evaluating imbalanced learning algorithms for AIDD in a plug-and-play manner. It supports research and development for solutions to real-world drug discovery problems such as molecular modeling, drug-target interaction and retrosynthesis.}
    \vspace*{-\baselineskip}
    \label{fig:overview}
\end{figure*}

Deep imbalanced learning~\cite{he2009learning,krawczyk2016learning}, a paradigm aiming to address the aforementioned challenges, remains relatively understudied~\cite{johnson2019survey} despite its importance and practicality for real-world applications. In the scenario of extreme imbalance ratio and large number of classes, it is closely connected to long-tailed recognition~\cite{LT_Survey,zhang2021bag} and subpopulation shift~\cite{bickel2007discriminative,barocas2016big}. However, recent advances mostly focus on Computer Vision (CV) tasks with convolutional neural networks~\cite{huang2019deep,liu2019large,ClassBalanceFocal,jamal2020rethinking,zhang2021distribution}, where state-of-the-art performance has become relatively saturated~\cite{LT_Survey,zhang2021test}. To call attention to the prevalent imbalance issue in the pharmaceutical domain and drive algorithmic innovation for real-world problems, we fill the gap by presenting a comprehensive benchmark for Deep \textbf{Im}balanced Learning in AI-aided \textbf{Drug} Discovery, ImDrug for brevity.

\textbf{Present work.} With ImDrug\footnote{Source code and datasets are available at: \url{https://github.com/DrugLT/ImDrug}.}, we provide a platform to systematically implement and evaluate deep imbalanced learning across a whole spectrum of therapeutics, covering 11 AI-ready datasets and 54 learning tasks (Fig.~\ref{fig:overview}). Inspired by the recent progress and literature on long-tailed visual recognition~\cite{LT_Survey}, on top of 4 classical machine learning algorithms, we benchmark 10 competitive deep learning baselines which can be divided into 3 categories: class re-balancing, information augmentation and module improvement. In addition to the imbalanced and long-tailed classification, ImDrug also incorporates 2 realistic but more challenging imbalanced settings, namely open long-tailed classification~\cite{liu2019large} and imbalanced regression~\cite{yang2021delving}. These settings require intrinsically different treatments of data and label distribution, for which our benchmark sheds light on future innovation of effective algorithms.

The canonical data splitting of imbalanced learning, namely the standard split in ImDrug (Fig.~\ref{fig:overview}), is to randomly sample a training set while retaining a \emph{balanced} test set to ensure equality of all classes~\cite{LT_Survey,zeng2022imgcl}. However, for highly imbalanced (long-tailed) datasets, such splitting severely restricts the size of the test set by forcing the sample number of majority classes to equal that of minority classes. To improve the quality of evaluation statistics, we introduce 5 imbalanced data split functions following the best practice in chemical sciences to cover more realistic out-of-distribution scenarios, which allows for significantly larger test sets. We argue that commonly reported metrics such as accuracy and AUROC can be over-optimistic under imbalanced distributions. Accordingly, two imbalanced learning metrics, \emph{balanced accuracy}~\cite{sokolova2006beyond,brodersen2010balanced} and a novel measure \emph{balanced F1}, are proposed to provide better insight of the evaluations due to their robustness to test distributions.

In summary, we make three key \textbf{contributions} in this work:
\begin{itemize}[leftmargin=10pt]
\setlength\itemsep{0em}
    \item \emph{Systematic and focused study at the intersection of AIDD and deep imbalanced learning.} We identify data imbalance as a major challenge in AIDD. In response, we build ImDrug, to our best knowledge the first comprehensive platform for benchmarking up-to-date deep imbalanced learning methods across datasets and tasks spanning the drug discovery lifecycle. Our analysis shows that there is ample room for further algorithmic improvement.
    \item \emph{Accessible framework for configuring customized datasets and algorithms.} We provide a Python package and unified configuration tools to allow for curation of new datasets and imbalanced learning algorithms, with a scalable modular and hierarchical design.
    \item \emph{Extended scope and novel metrics.} In addition to the conventional imbalanced and long-tailed classification, ImDrug encompasses 2 practically important and challenging settings with an empirical study of the corresponding specialized algorithms. It features up-to-date chemical reaction datasets with multi-class labels to provide benchmarks of long-tailed learning methods. Moreover, we identify the limitation of standard data splits and evaluate baselines on various imbalanced splits via two robust metrics, balanced accuracy and newly proposed balanced F1.
\end{itemize}

%% file: Sections-TPAMI/2-Related_work.tex
\section{Related Work}\label{sec:related_work}
ImDrug is the first systematic and unified platform targeting deep imbalanced learning in AIDD. To date, the researches on imbalanced learning and AIDD follow two separate lines with little overlap, which we briefly review in this section.

\textbf{Relation to imbalanced learning literature and benchmarks.}
Imbalanced learning, or long-tailed learning in extreme cases (Table~\ref{table:settings}), remains a long-standing problem in machine learning~\cite{he2009learning,krawczyk2016learning}. 
% In the past decades, along with the rise of deep learning and neural networks, a surge of datasets and benchmarks have been constructed on deep imbalanced learning in Computer Vision (CV)~\cite{LT_Survey,zhang2021bag,huang2019deep,liu2019large,LDAM,INaturalist,DBLP:conf/eccv/WuH0WL20,garcin2021pl}, Natural Language Processing (NLP)~\cite{yang2021delving,DBLP:conf/emnlp/RajpurkarZLL16,DBLP:conf/acl/RajpurkarJL18,DBLP:conf/emnlp/DasigiLMSG19,wang2019superglue} and social network domains~\cite{tang2008arnetminer,mohammadrezaei2018identifying,hu2020open}. To account for different data structures in each domain, these benchmarks employ distinct backbone architectures as feature extractors, namely CNN~\cite{he2016deep} for CV, Transformer~\cite{vaswani2017attention} for NLP and GNN~\cite{scarselli2008graph} for graphs. In ImDrug, due to the multi-modal nature of molecular data, we experiment with MLP, Transformer and GCN~\cite{kipf2017semisupervised} for numeric ($e.g.$ Morgan Fingerprint~\cite{DBLP:journals/jcisd/RogersH10}), sequence ($e.g.$ SMILES~\citep{weininger1988smiles}) and graph feature representations respectively.
% On top of various model architectures for distinct modalities, as the state-of-the-art (SOTA) technologies advance, several literature reviews have applied different taxonomies for contemporary imbalanced learning methods. In the classical machine learning era,
\citet{he2009learning} partition the imbalanced learning methods into 5 categories, including sampling, cost-sensitive, kernel-based, active learning and other methods. Later, \citet{krawczyk2016learning} distinguishes 3 general approaches to learning from imbalanced data: \emph{data-level methods} by re-balancing samples, \emph{algorithm-level methods} by algorithmic bias alleviation and adaptation to skewed distribution, and \emph{hybrid methods} that combine the advantages of two previous groups. Most recently, by going through the bulk of deep long-tailed recognition methods published since 2016, ~\citet{LT_Survey} make a new taxonomy consisting of class re-balancing, information augmentation and module improvement with 9 sub-classes. In this work, we implement and benchmark 10 representative imbalanced learning methods following~\cite{LT_Survey} to better reflect the up-to-date progress in this field.
% following~\citet{LT_Survey}

\textbf{Relation to AIDD repositories.} For drug discovery, there exists a myriad of large-scale databases with different focuses. ChEMBL~\cite{mendez2019chembl,davies2015chembl} contains over 2.1 million compounds and 18.6 million bioassay records of their activities. BindingDB~\cite{gilson2016bindingdb} provides a collection of binding affinity data between small molecules and proteins. USPTO~\cite{lowe2012extraction} curates nearly 2 million reaction data mined from US patents. These biorepositories are naturally imbalanced and serve as valuable resources for ImDrug. However, they require careful pre-processing to be amenable to deep learning models, due to inconsistent data representations and noisy labels.

Based on the preceding raw databases, several AI-ready datasets and benchmarks have been built recently. MoleculeNet~\cite{wu2018moleculenet} provides among the first large-scale repositories for molecular machine learning and quantitative structure-activity relationship (QSAR)~\cite{dudek2006computational} modeling, with benchmarks for classical machine learning and graph-based models. The authors briefly discuss the class imbalance in the datasets and suggested using AUROC and AUPRC for evaluation. The settings were largely inherited in follow-up works such as TDC~\cite{huang2021therapeutics}, which extends the scope of modalities and learning tasks to full coverage of therapeutics pipelines. However, most of the existing benchmarks like MolecuNet and TDC offer binary classification or regression prediction tasks only. 

In contrast, our work builds upon the tiered design of TDC and introduces extra datasets for virtual screening~\cite{2022arXiv220109637J} and chemical reactions. It features a series of up-to-date USPTO datasets~\cite{schwaller2021mapping,lu2022unified,lowe2012extraction,huang2021therapeutics,liu2017retrosynthetic} to support \emph{multi-class prediction} of reaction type, template, catalyst and yield.
% as opposed to the \emph{binary classification tasks} in MoleculeNet and TDC. 
Moreover, ImDrug carries out a systematic and focused study of deep imbalanced learning in AIDD with comprehensive settings and newly proposed metrics. We highlight the observations that data imbalance poses a key obstacle for developing trustworthy AIDD solutions, justifying the need and opportunity for algorithmic/data innovation at the intersection of AIDD and deep imbalanced learning.

%% file: Sections-TPAMI/3-Overview.tex
\section{Overview}
\label{sec:overview}

\subsection{High-Level Infrastructure}
The design principle of ImDrug is to provide a platform with full coverage of the algorithm development lifecycle in AIDD. From a high-level view, the five-stage ImDrug pipeline illustrated in Fig.~\ref{fig:overview} consists of a superposition of efforts in two orthogonal dimensions: \emph{curation of AIDD datasets/learning tasks} and \emph{benchmarking imbalanced learning algorithms}. 

\subsubsection{Tiered Design of Datasets \& Learning Tasks}
\label{sec:3.1.1}

We employ a three-level hierarchical structure for organizing datasets and learning tasks. At the top level, inspired by TDC~\cite{huang2021therapeutics}, we categorize 54 learning tasks into three prediction problems:

\begin{itemize}[leftmargin=10pt]
\setlength\itemsep{0em}
    \item \textbf{Single-instance prediction:}  Predictions based on individual biomedical entities, including 44 tasks on small molecules.
    \item \textbf{Multi-instance prediction:} Predictions based on multiple heterogeneous entities, including 2 tasks on virtual screening.
    \item \textbf{Hybrid prediction:} Predictions based on a set of homogeneous entities, which can be either aggregated ($e.g.$, string or graph concatenation) to perform single-instance prediction, or separately encoded to apply multi-instance prediction, including 7 tasks on chemical reactions and 1 on drug-drug interaction.
\end{itemize}

At the middle level, each prediction problem above is assigned with a collection of datasets:

\begin{itemize}[leftmargin=10pt]
\setlength\itemsep{0em}
    \item 4 compound datasets for prediction of the ADMET (absorption, distribution, metabolism, excretion and toxicity), bioactivity and quantum chemical properties.
    \item 1 drug-target interaction dataset for prediction of the protein-ligand binding affinity.
    \item 5 reaction datasets for prediction of the chemical reaction type, template, catalyst and yield. 1 dataset for drug-drug interaction prediction.
\end{itemize}

A detailed summary of the 11 datasets' statistics is shown in Table~\ref{table:datasets}. At the bottom level, we provide data processing utilities involving 6 split functions and 23 featurizers customized for compound and protein. Besides the standard splitting with a balanced test set, ImDrug offers 5 imbalanced data split functions to account for settings such as out-of-distribution (OOD) generalization~\cite{2022arXiv220109637J} and continual learning~\cite{delange2021continual}, with improved quality of evaluation statistics by enabling a significantly larger test set in the highly imbalanced scenario. Additionally, our proposed metrics in Sec. \ref{sec:evaluation_metrics} ensure fairness with a mathematical guarantee that testing on imbalanced splits is equivalent to evaluation on a balanced test set (Thm.~\ref{thm:balanced-f1}).

\input{Tables/Settings}
\input{Tables/Datasets}

\subsubsection{Imbalanced Learning Settings}

For benchmarking imbalanced learning algorithms, to start with, we define the imbalanced learning problem as follows. Consider $\{x_i, y_i\}_{i=1}^n$ as an imbalanced dataset consisting of $n$ sample-label pairs drawn from $K$ classes with $n=\sum_{k=1}^K n_k$, where $n_k$ denotes the number of data points of class $k$ and $\pi_k = n_k/n$ represents the label frequency of class $k$. Without loss of generality, we assume the classes are sorted by cardinality in decreasing order~\cite{LT_Survey}, $i.e.$, if $i<j$, then $n_i \ge n_j$, and $n_1 \gg n_k$. We denote by $n_1/n_K$ the imbalance ratio of the dataset.

Based on the setup above, we formulate 4 realistic and practically important imbalanced learning settings summarized in Table~\ref{table:settings}. Besides the regular imbalanced classification, which is usually measured by metrics like accuracy, F1 score, and AUROC, we observe that some AIDD datasets contain extremely large number of classes with high imbalance ratio. For instance, in terms of reaction type classification, USPTO-50k~\cite{liu2017retrosynthetic}, USPTO-500-MT~\cite{lu2022unified} and USPTO-1k-TPL~\cite{schwaller2021mapping} include 10, 500 and 1000 classes with imbalance ratio 22.6, 285.0 and 117.4 respectively. These datasets exhibit power-law scaling of label frequency distributions, which are attributed to the long-tailed (LT) classification protocol of ImDrug. Conventionally, evaluations on LT classification report statistics for the head, middle, tail, and overall classes separately.

Additionally, open long-tailed classification (Open LT)~\cite{liu2019large} as the third setting, introduces more challenging problems by demanding the algorithms to not only strike a balance between majority and minority classes, but also be able to generalize to unseen (open) classes at test time. This scenario is closely connected to the out-of-distribution detection in AIDD~\cite{2022arXiv220109637J} and cost-sensitive online classification~\cite{wang2013cost}. Besides the shift of class labels implemented in ImDrug, examples of open-set distribution shifts in drug discovery include domain shift of molecular size and scaffold, cold-start of new protein targets, and a temporal shift of patented chemical reactions as in Fig.~\ref{fig:overview}.

The last setting, imbalanced regression, calls for attention of the data imbalance issue in continuous label space~\cite{yang2021delving}. As opposed to the categorical target space of classification tasks, the continuous label spectrum inherently enforces a meaningful distance between targets, which has implications for how one should interpret data imbalance. This setup is not as well explored and extends the scope of ImDrug to a new regime of real-world AIDD applications, such as the prediction of quantum chemical properties~\cite{qm9_1,qm9_2}, protein-ligand binding affinity~\cite{2022arXiv220109637J} and reaction yields~\cite{lu2022unified}.

\subsection{Evaluation Metrics}\label{sec:evaluation_metrics}

For conventional imbalanced classification and regression in domains like Computer Vision, to ensure equal weighting of all classes, a balanced test set is selected~\cite{LT_Survey,yang2021delving}. However, in highly imbalanced datasets ($i.e.$, high imbalance ratio), such protocol severely constrains the size of the test set. Considering the Central Limit Theorem and the fact that the uncertainty of the point-estimated mean scales with $1/\sqrt{n}$ ($n$ being the number of samples), ideally we would like to have a much larger test set, $e.g.$, $10\text{-}20\%$ of the whole dataset to achieve satisfactory statistical significance at evaluation.

However, reporting metrics like accuracy and F1 score on an imbalanced test set can be problematic in two ways~\cite{brodersen2010balanced}: first, it does not allow for the derivation of meaningful
confidence intervals; second, it leads to an optimistic estimate in presence of a biased classifier. To avoid these pitfalls, we adopt the balanced accuracy~\cite{sokolova2006beyond,brodersen2010balanced} and propose a novel balanced F1 score which can be used as an unbiased measure on any imbalanced test sets. To formalize our proposed metric, we define the balanced multi-class precision analogous to the calibrated precision introduced in~\cite{siblini2020master}:

\begin{definition}\label{def:balanced_metrics}
Given a test set $\{x_i, y_i\}_{i=1}^n$ of $n$ samples drawn from $K$ classes and the corresponding predicted labels $\{\widehat{y}_i\}_{i=1}^n$, the balanced accuracy/recall (BA) and the balanced precision for class $k$ ($\text{BP}_k$) are defined as:
\begin{align}
    \text{BA} &\coloneqq
    \frac{1}{K}\sum_{k=1}^K Rec_k =
    \frac{1}{K}\sum_{k=1}^{K}\frac{\sum\limits_{i=1}^{n}\mathbbm{1}(y_i, \widehat{y}_i=k)}{\sum\limits_{i=1}^n\mathbbm{1}(y_i=k)}\label{eqn:balanced-acc}\\
    \text{BP}_k &\coloneqq 
    \frac{\sum\limits_{i=1}^n\mathbbm{1}(y_i, \widehat{y}_i=k)}{\sum\limits_{i=1}^n\mathbbm{1}(y_i, \widehat{y}_i=k) + \sum\limits_{j\ne k}\sum\limits_{i=1}^n\pi_{jk}\mathbbm{1}(y_i=j, \widehat{y}_i=k)}\ ,\label{eqn:balanced-prec}
\end{align}

\noindent where $Rec_k$ stands for the recall for class $k$ and the calibration factor $\pi_{jk} = \sum_{i=1}^{n}\mathbbm{1}(y_i=k)/\sum_{i=1}^{n}\mathbbm{1}(y_i=j) = n_k/n_j$ is the ratio between the label frequencies of class $k$ and $j$. When $\pi_{jk}\equiv 1$, $i.e.$, the dataset is perfectly balanced, Eqn.~\ref{eqn:balanced-prec} recovers the conventional one-vs-all precision for class $k$.
\end{definition}

Note that, unlike the balanced precision, the balanced recall for class $k$ is equivalent to the conventional one-vs-all recall for the same class, since its calculation does not involve the support of other classes, which induces the following theorem:

\begin{theorem}\label{thm:balanced-f1}
Let $Rec_k$ and $\text{BP}_k$ denote the recall and balanced precision for class $k$. Given a trained predictor, the evaluated balanced F1 score

\begin{equation}
    \text{Balanced-F1} \coloneqq \frac{1}{K}\sum_{k=1}^{K}\frac{2\times Rec_k\times \text{BP}_k}{Rec_k + \text{BP}_k} 
\end{equation}

\noindent is invariant to label distribution shift on any test set, assuming samples of each class are drawn i.i.d from a fixed distribution. 
\end{theorem}

\begin{proof}
See Appendix B.
% ~\ref{sec:appendix:proof}.
\end{proof}

\subsection{Imbalanced Learning Baselines} 

In this section, we provide a brief review of benchmarked baseline methods for the 4 imbalanced learning protocols summarized in Table~\ref{table:settings}. A comprehensive description can be found in the Appendix.

\subsubsection{Baselines of Classical Machine Learning}
4 Baselines including Random Forest, Decision Tree, K-nearest Means and SMOTE~\cite{chawla2002smote}, a method tailored for imbalanced learning are considered. These ML-based models/algorithms are implemented without any neural network architectures, and found to consistently underperform on ImDrug datasets featuring large size and many classes (Table~\ref{table:datasets}). Therefore they are compared to other deep imbalanced learning baselines in our main experiments only (Table~\ref{table:overall}).

\subsubsection{Baselines of Imbalanced \& Long-Tailed Classification}\label{sec:classification_baselines}

Besides the vanilla baseline trained by the softmax cross-entropy loss, following the taxonomy proposed by~\citet{LT_Survey}, we consider 10 baselines in 3 families of conventional deep imbalanced/long-tailed classification methods: 4 for class re-balancing
% \footnote{The explicit expressions of the cost-sensitive loss functions are provided in Appendix.} 
% illustrated in Table~\ref{table:loss_functions}
(\textbf{Cost-Sensitive Loss (CS)}~\cite{CostSensitiveCE}, \textbf{Class-Balanced Loss (CB\_F)}~\cite{ClassBalanceFocal}, \textbf{Balanced Softmax (BS)}~\cite{BalancedSoftmaxCE}, \textbf{Influence-Balanced Loss (IB)}~\cite{park2021influence}), 3 for information augmentation (\textbf{Mixp}~\cite{zhang2018mixup}, \textbf{Remix}~\cite{chou2020remix}, \textbf{DiVE}~\cite{DiVE}) and 3 for module improvement (\textbf{CDT}~\cite{CDT}, \textbf{Decoupling}~\cite{kang2020decoupling}, \textbf{BBN}~\cite{zhou2020bbn}). 

For implementation, we employ a modular design based on~\cite{zhang2021bag} and choose the baselines which can be realized as an individual “trick” in terms of loss functions (class re-balancing) summarized in Table~\ref{table:loss_functions}, data samplers (module improvement) or instance combiners (information augmentation). The selection of baselines reflects a notable scientific progress and paradigm shift of imbalanced learning algorithms from 2002 to date. 

\begin{table}
\centering
\caption{\textbf{Summary of re-balancing losses.} $z$ and $p$ denote the predicted logits and the softmax probability of the sample $x$, with subscript $y$ being the corresponding class label. $n_y$ and $\pi_y$ indicate the sample number and label frequency of the class $y$. $\gamma$ and $\beta$ are loss-related hyperparameters.}
\renewcommand{\arraystretch}{1.5}
\label{table:loss_functions}
\setlength{\tabcolsep}{4pt}
\begin{adjustbox}{max width=\columnwidth}
\begin{tabular}{ll}
\toprule
Losses & Formulation \\ \midrule
Softmax loss (vanilla) & $\mathcal{L}_{CE} = -\log(p_y)$\\
Cost-sensitive loss~\cite{CostSensitiveCE} & $\mathcal{L}_{CS} =  -\frac{1}{\pi_y}\log(p_y)$\\
Class-balanced focal loss~\cite{ClassBalanceFocal} & $\mathcal{L}_{CB\_F} = -\frac{1-\gamma}{1-\gamma^{n_y}}(1-p_y)^{\beta}\log(p_y)$\\
Class-balanced cross-entropy loss~\cite{ClassBalanceFocal} & $\mathcal{L}_{CB\_CE} = -\frac{1-\gamma}{1-\gamma^{n_y}}\log(p_y) $\\
Balanced softmax loss~\cite{BalancedSoftmaxCE} & $\mathcal{L}_{BS} = -\log\left(\frac{\pi_y\exp(z_y)}{\sum_j\pi_j\exp(z_j)}\right)$\\
Influence-balanced loss~\cite{park2021influence} & $\mathcal{L}_{IB} = -\frac{1}{\pi_y||\nabla\log(p_y)||_1}\log(p_y)$\\
\bottomrule
\end{tabular}
\end{adjustbox}
\end{table}

\subsubsection{Baselines of Open LT \& Imbalanced Regression}

\textbf{Open LT} can be regarded as a variant of conventional long-tailed classification with unseen and out-of-distribution tail classes in the test set. In principle, all aforementioned baselines for imbalanced classification can be seamlessly transferred to this setting. However, the extra challenge posed by the protocol requires targeted remedies to OOD detection to achieve state-of-the-art performance. Hence ImDrug includes 2 additional baselines for Open LT: \textbf{OLTR}~\cite{liu2019large} and \textbf{IEM}~\cite{zhu2020inflated}.

\textbf{Imbalanced Regression} can be reduced to conventional imbalanced classification by naively dividing the continuous label space into multiple consecutive bins as classes. From this perspective, many baselines in Sec.~\ref{sec:classification_baselines} can be adapted by replacing the classification head with a regression head. However, such trivial transformation undermines the intrinsic topology of labels induced by the Euclidean distance, leading to sub-optimal performance~\cite{yang2021delving}. In ImDrug, we experiment with 3 extra baselines specialized at solving this particular challenge: \textbf{Focal-R}~\cite{yang2021delving}, \textbf{LDS}~\cite{yang2021delving} and \textbf{FDS}~\cite{yang2021delving}.

%% file: Tables/Settings.tex
\begin{table*}[]
\centering
\caption{\textbf{Comparison between the 4 imbalanced learning settings investigated in ImDrug.}}

\label{table:settings}
\setlength{\tabcolsep}{4pt}
\begin{adjustbox}{max width=\textwidth}
\begin{tabular}{l|ccclc}
\toprule
Setting               & Label Type                                                       & Imbalance Ratio       
                                           & \# Classes                    & Evaluation Metrics                         & Open Class at Test Time \\ \midrule

Imbalanced Classification & 
Categorical               &
Low                      &                    
$< 10$                    & 
Acc, F1, AUROC   & \xmark \\

Long-Tailed Classification & 
Categorical                &
High                       &                    
$\ge 10$                   & 
Acc, F1, AUROC (+head, middle, tail)                         & \xmark \\

Open Long-Tailed Classification & 
Categorical                &
High                       &                    
$\ge 10$                   & 
Acc, F1, AUROC (+head, middle, tail, open)                      & \cmark \\

Imbalanced Regression      & 
Continuous                 &
N/A                        &                    
N/A                        & 
MSE, MAE (+head, middle, tail)                                &
N/A \\
\bottomrule

\end{tabular}
\vspace*{-0.15in}
\end{adjustbox}
\vspace*{-0.15in}
\end{table*}

%% file: Tables/Datasets.tex
\begin{table*}[]
\caption{\textbf{Statistics of the ImDrug datasets.}}\label{table:datasets}
\begin{adjustbox}{max width=\textwidth}
\begin{tabular}{ll|cccccc}
\toprule
Dataset                & Learning Tasks                                                                                               & \multicolumn{1}{l}{Size} & \multicolumn{1}{l}{\# Classes} & Imbalance Ratio & Unit   & Feature   & Recommended Settings                        \\ \midrule
ImDrug.BBB\_Martins~\cite{huang2021therapeutics}    & single\_pred.ADMET                                                                                           & 1975                     & 2    &  3.24                          & -      & Seq/Graph & Imbalanced Classification            \\
ImDrug.Tox21~\cite{huang2021therapeutics}           & single\_pred.ADMET                                                                                           & 7831                     & 2    & 22.88                         & -      & Seq/Graph & Imbalanced Classification            \\
ImDrug.HIV~\cite{huang2021therapeutics}             & single\_pred.BioAct                                                                                             & 41,127                   & 2  & 27.50                             & -      & Seq/Graph & Imbalanced Classification            \\
ImDrug.QM9~\cite{qm9_1,qm9_2}             & single\_pred.QM                                                                                              & 133,885                  & - & -          & GHz/D/$^2_0$/$^3_0$ & Coulomb   & Imbalanced Regression                \\
ImDrug.SBAP~\cite{2022arXiv220109637J}        & multi\_pred.DTI                                                                                              & 32,140                   & 2 & 36.77                             & nM       & Seq/Graph & Imbalanced Classification/Regression \\
ImDrug.USPTO-Catalyst~\cite{huang2021therapeutics} & hybrid\_pred.Catalyst                                                                                        & 721,799                  & 888  &  3975.9                       & -      & Seq/Graph & LT Classification/Open LT            \\

ImDrug.USPTO-50K~\cite{liu2017retrosynthetic}       & hybrid\_pred.ReactType                                                                                       & 50,016                   & 10   & 65.78                          & -      & Seq/Graph & LT Classification/Open LT            \\
ImDrug.USPTO-500-MT~\cite{lu2022unified}     &  hybrid\_pred.ReactType  & 143,535                  & 500    & 285.06                            & -      & Seq/Graph & LT Classification/Open LT            \\
ImDrug.USPTO-500-MT~\cite{lu2022unified}      &  hybrid\_pred.Catalyst  & 143,535                & 27,759   & -                          & -    & Seq/Graph  & LT Classification/Open LT          \\
ImDrug.USPTO-500-MT~\cite{lu2022unified}      &  hybrid\_pred.Yields  & 143,535               & -   & -                           & \%      & Seq/Graph & Imbalanced Regression            \\
ImDrug.USPTO-1K-TPL~\cite{schwaller2021mapping}    & hybrid\_pred.ReactType                                                                                       & 445,115                & 1000   & 110.86                       & -      & Seq/Graph & LT Classification/Open LT            \\
ImDrug.USPTO-Yields~\cite{huang2021therapeutics}    & hybrid\_pred.Yields                                                                                          & 853,638                  & - & -          & \%     & Seq/Graph & Imbalanced Regression                \\

ImDrug.DrugBank\cite{huang2021therapeutics}        & hybrid\_pred.DDI                                                                                             & 191,808                  & 86 & 10125                            & -      & Seq/Graph & LT Classification/Open LT  \\ \bottomrule        
\end{tabular}
\end{adjustbox}
\end{table*}

%% file: Sections-TPAMI/4-Experiments.tex
\newcommand{\eat}[1]{}
\section{Experiments}
\label{sec:experiments}

In experiments, we benchmark 11 baselines for conventional imbalanced \& long-tailed classification (Sec.~\ref{sec:classification_baselines}), as well as 5 additional baselines tailored for open LT and imbalanced regression, by reporting the 2 proposed balanced accuracy and balanced F1 measures along with the conventional AUROC. Other metrics for imbalanced learning such as AUPRC and Weighted-F1 are also reported in Appendix H. For hybrid prediction tasks/datasets (Table~\ref{table:datasets}), unless otherwise specified, the training mode is single-instance prediction by default.
All average performance with standard deviations is evaluated over 3 random seeds. 

In ImDrug, due to the multi-modal nature of molecular data, we experiment with MLP,
Transformer~\cite{vaswani2017attention}, and GCN~\cite{kipf2017semisupervised} for numeric- ($e.g.$, Morgan Fingerprint~\cite{DBLP:journals/jcisd/RogersH10}), sequence- ($e.g.$, SMILES~\citep{weininger1988smiles})
and graph-based feature representations respectively. We provide ablation study of backbone models in Table~\ref{table:backbones}. GCN~\cite{kipf2017semisupervised} is chosen as the default backbone for the rest of experiments due to its stable and superior performance. The input graph encoding is implemented via the DGL library~\cite{wang2019dgl}.
% Due to space limitations, additional empirical studies such as ablations of the backbone models are provided in Appendix.

\input{Tables-new/table-1-new}

\input{Tables-new/table-3-new}

\begin{figure}[b]
    \centering
    \includegraphics[width=\columnwidth]{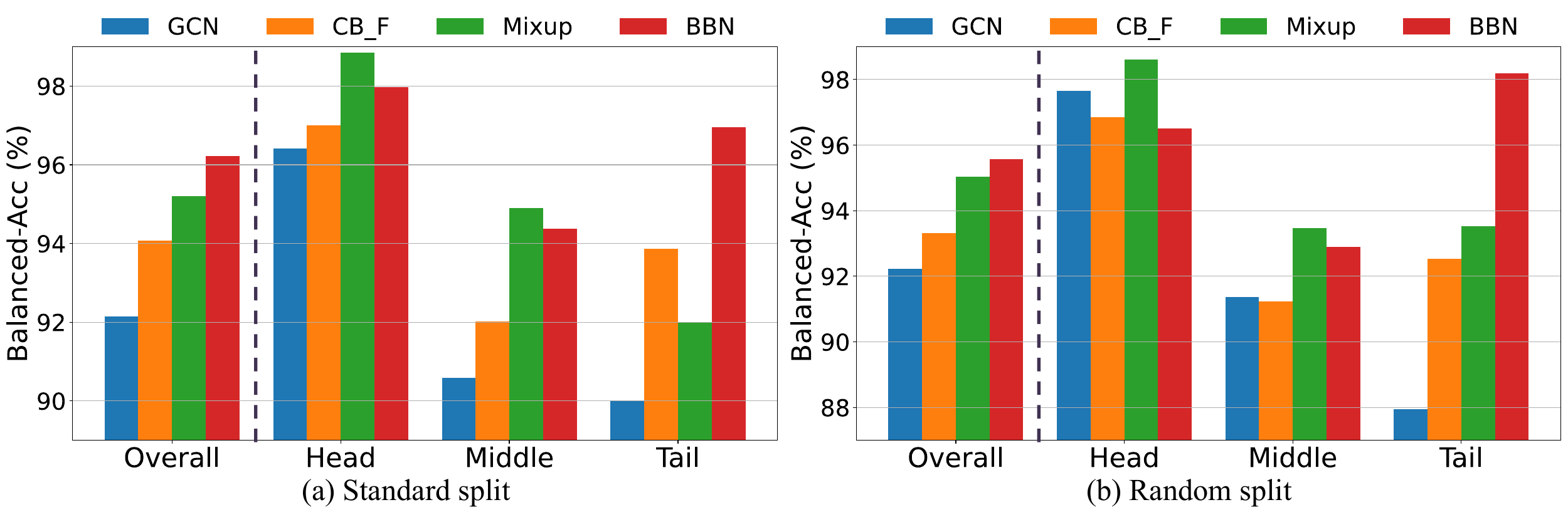}
    \caption{\textbf{Performance of single-instance prediction for the dataset USPTO-50K of overall, head, middle and tail under standard and random splits.} Balanced accuracy for 4 baseline methods on different class subsets is shown.}
    \label{fig:head-mid-tail}
    \centering
\end{figure}

\subsection{Results on Imbalanced \& Long-Tailed Classification}
\label{sec:4.1}

\begin{figure}[b]
    \includegraphics[width=\columnwidth]{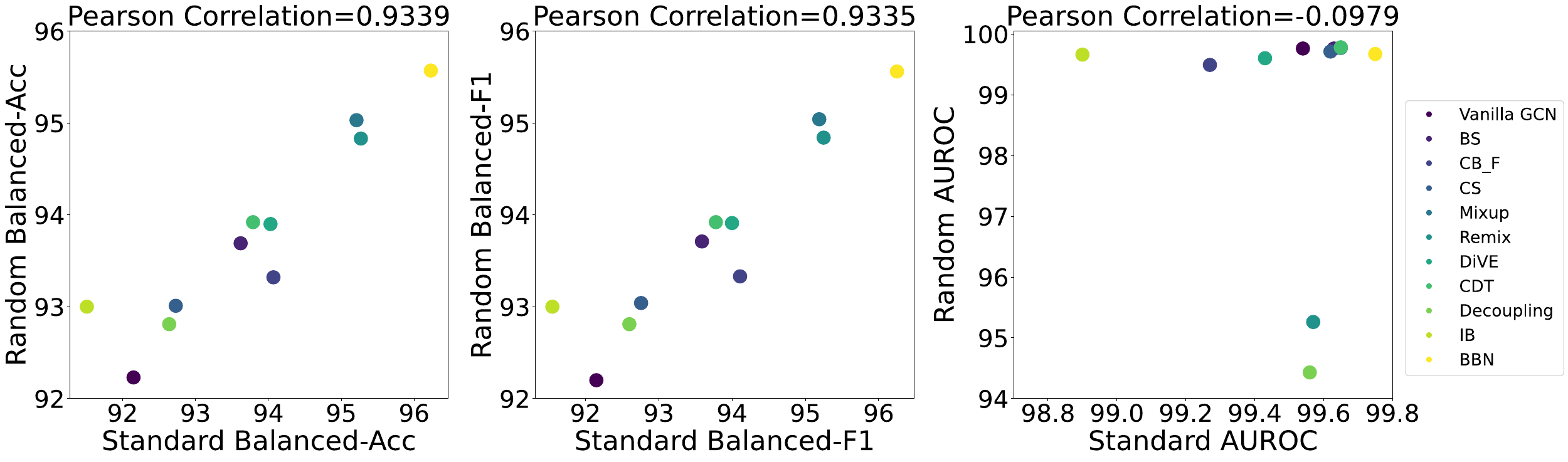}
    \caption{\textbf{Comparison of the baseline performance for standard and random splits on USPTO-50K under single-instance prediction}, in terms of balanced accuracy, balanced F1 and AUROC. The 2 proposed balanced metrics exhibit a good correlation between the two splits (0.9339, 0.9335), which is significantly higher than AUROC (0.0979).}
    \label{fig:compare}
\end{figure}

Table~\ref{table:overall} reports the average performance of imbalanced learning baselines on 2 binary ($aka$., imbalanced) classification and 2 multi-class ($aka$., long-tailed) classification datasets. For the 2 proposed balanced metrics, illustrated in Fig.~\ref{fig:compare}, we observe a high correlation between the performance on balanced standard split and imbalanced random split, which is in agreement with the theory (Thm.~\ref{thm:balanced-f1}) that both metrics are insensitive to label distributions. Moreover, almost all baselines outperform the vanilla GCN and classical machine learning methods, demonstrating the effectiveness of deep imbalanced learning. However, we observe that the SOTA models do not work well consistently across the ImDrug datasets and tasks. For example, the information augmentation algorithms are most effective on USPTO-50K but subpar on the others. Whereas the class-rebalancing approaches are highly competitive on HIV and DrugBank. Overall, module improvement methods such as BBN exhibit the most consistent and superior performance, which aligns with the empirical findings by~\citet{LT_Survey} in the CV domain and explains the paradigm shift of deep imbalanced learning towards this direction in recent years~\cite{LT_Survey}. 

Besides the results on all classes, Fig.~\ref{fig:head-mid-tail} breaks down the performance of 4 baselines on USPTO-50K into the head, middle, and tail classes. Evidently, effective imbalanced learning methods like BBN and Remix increase the overall performance by consistently improving on all class subsets, with the most significant gain on the tail classes.

\input{Tables-new/table-2-new}

By comparing the results on the three evaluation metrics, we observe consistent rankings, especially between the two balanced metrics (also shown in Fig.~\ref{fig:corr}). Moreover, on these highly-imbalanced AIDD datasets, AUROC offers over-optimistic results and is typically insensitive to different baselines. In contrast, the proposed balanced metrics more effectively highlight the challenges posed by the imbalanced settings, by significantly up-weighting the contribution from the minority classes and hard samples. The fact that even the SOTA algorithms fail to reach $80\%$ accuracy for binary classification on HIV and $24\%$ accuracy for 888-way classification on USPTO-Catalyst
\eat{and $50\%$ accuracy for 10-way classification on USPTO-50K}, indicates there is ample room for methodological innovation and improvement for real-world AIDD problems.

\subsection{Results on Open LT \& Imbalanced Regression}
% We postulate the reason that 
% % since they pay close attention to the open-set classification. 
% OLTR and IEM can both be performed with embeddings from the arbitrary baseline, the BBN baseline outperforms the baseline by up to 9\% on average, which hence derives the further development requirements for algorithms simultaneously focusing on tail-set and open-set classification.

\begin{table}[b]
\caption{\textbf{Results on imbalanced regression under standard split.} We report the average performance on SBAP-Reg (multi-instance prediction) and QM9 (single-instance prediction) in terms of mean MSE and MAE with standard deviation in this setting.}
\label{table:ImbalancedRegression}
\centering 
\small
\setlength{\tabcolsep}{3pt}
\scalebox{0.92}{
\begin{tabular}{c|cc|cc}
\toprule
             & \multicolumn{2}{c|}{\textbf{SBAP-Reg}}                      & \multicolumn{2}{c}{\textbf{QM9}}                           \\ \cmidrule{2-5} 
             & \multicolumn{1}{c}{MSE} & \multicolumn{1}{c|}{MAE} & \multicolumn{1}{c}{MSE} & \multicolumn{1}{c}{MAE} \\ \midrule
Vanilla GCN & $1.66_{\pm1.03}$           & $0.92_{\pm0.25}$            & $76.17_{\pm42.82}$        & $6.57_{\pm1.92}$          \\
Mixup        & $\underline{1.10_{\pm0.51}}$           & $\underline{0.78_{\pm0.17}}$            &  $82.35_{\pm36.86}$       & $6.59_{\pm1.46}$                        \\
BBN          & $1.52_{\pm1.57}$           & $0.89_{\pm0.38}$            & $\underline{59.42_{\pm14.39}}$        & $6.02_{\pm0.77}$          \\ \midrule
Focal-R       & $1.38_{\pm1.08}$           & $0.84_{\pm0.31}$            & $94.29_{\pm57.38}$        & $7.23_{\pm2.65}$          \\
LDS          & $1.36_{\pm1.16}$           & $\underline{0.78_{\pm0.26}}$            & $\boldsymbol{16.01_{\pm6.58}}$        & $\boldsymbol{2.73_{\pm0.03}}$           \\
FDS          & $\boldsymbol{0.59_{\pm0.09}}$           & $\boldsymbol{0.54_{\pm0.02}}$            & $60.81_{\pm33.52}$        & $\underline{5.63_{\pm1.53}}$           \\\bottomrule
\end{tabular}
}
\end{table}

For Open LT setting, we further conduct experiments on 2 multi-class classification datasets, and report the mean performance with standard deviation under random split in Table~\ref{table:openLT}. 
Observations are two-fold. 
First, deep imbalanced learning methods consistently outperform the vanilla GCN baseline in this setting, which demonstrates their effectiveness. 
Second, specifically designed algorithms for Open LT, such as OLTR, outperform methods for conventional imbalanced learning. Moreover, the fact that OLTR and IEM can directly operate on the processed embeddings of conventional baselines like BBN makes the combination of these tricks synergistic, which aligns with the observations in~\cite{LT_Survey,zhang2021bag}. Indeed, OLTR+BBN outperforms the BBN baseline by up to 3.40\% on average, which highlights the benefit of exploring the combinatorial design space of baselines across various imbalanced learning settings.

For imbalanced regression, we report the mean results with standard deviations in terms of MSE and MAE on 2 imbalanced regression datasets in Table~\ref{table:ImbalancedRegression}. 
We find that algorithms tailored for imbalanced regression outperform all conventional imbalanced classification methods. 
In particular, FDS and LDS stand out by beating all other competing methods in this experimental setting.
However, despite their effectiveness, the improvements are still marginal on datasets like QM9, which calls for more effective solutions targeting imbalanced regression.

\subsection{Evaluations of the Specific Designs}
Besides the general findings on the 4 imbalanced learning settings, in this section, we assess the effectiveness of the specific benchmarking utilities introduced by ImDrug. With experiments and analysis, we aim to address the following two core questions:
\begin{itemize}[leftmargin=10pt]
\setlength\itemsep{0em}
\item\textbf{Q1:} Can ImDrug provide meaningful proxies of the real-world challenges in AIDD?  (Sec.~\ref{sec:ood_splits})
\item\textbf{Q2:} Can ImDrug address the loss of efficacy of conventional metrics ($e.g.$, regular accuracy) on imbalanced test sets? (Sec.~\ref{sec:metrics_correlation})
\end{itemize}

\subsubsection{Observations on the OOD Splits}\label{sec:ood_splits}
\begin{figure}[b]
    \centering
    \includegraphics[width=\columnwidth]{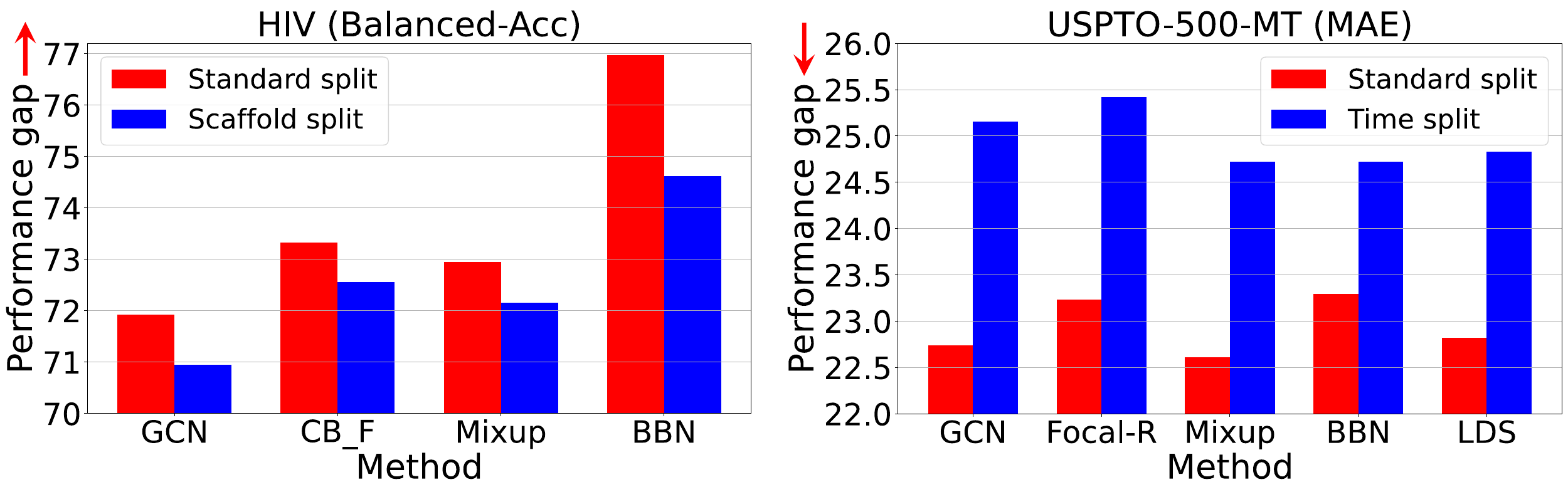}
    \caption{\textbf{Evaluations of baselines for binary classification} on HIV with scaffold split, and imbalanced regression on USPTO-500-MT with temporal split. Compared to the in-distribution standard split, the OOD splits on average induce performance gaps of $1.22\%\downarrow$ in balanced accuracy and $2.03\%\uparrow$ in MAE, respectively. The arrows indicate the direction of better performance.}
    \label{fig:split}
\end{figure}
One of the most pressing real-world challenges in AIDD is domain/distribution shifts~\cite{huang2021therapeutics,2022arXiv220109637J}, which is prevalent in QSAR problems such as small-to-large molecule generalization for ADMET prediction, cold-start of protein targets ($e.g.$, unseen antigens of COVID-19) for drug-target interaction and predictions of unknown reaction types for retrosynthesis. Fortunately, as in Fig.~\ref{fig:overview}, ImDrug provides testbeds for these domain-specific challenges with 5 extra data split functions. 

Among them, the scaffold split poses harder challenges than random split for single-instance predictions by assigning the compounds with identical molecular scaffold~\cite{bemis1996properties} to the same train/valid/test sets, making the split datasets more structurally different. 
For USPTO repositories curated from chemical reactions with patent date information, temporal split mimics real-world drug discovery challenges by assigning all data published before a specific point in time to the training set and those patented later to the test set, thus enforcing OOD in the temporal dimension. We investigated these two OOD splits on HIV and USPTO-500-MT respectively. As shown in Fig.~\ref{fig:split}, both splits result in consistently degraded performance across all baselines, highlighting the need for more advanced AIDD solutions addressing OOD and label imbalance simultaneously.

\subsubsection{Observations on the Proposed Metrics}\label{sec:metrics_correlation}

\begin{figure}
    \centering
    \includegraphics[width=\columnwidth]{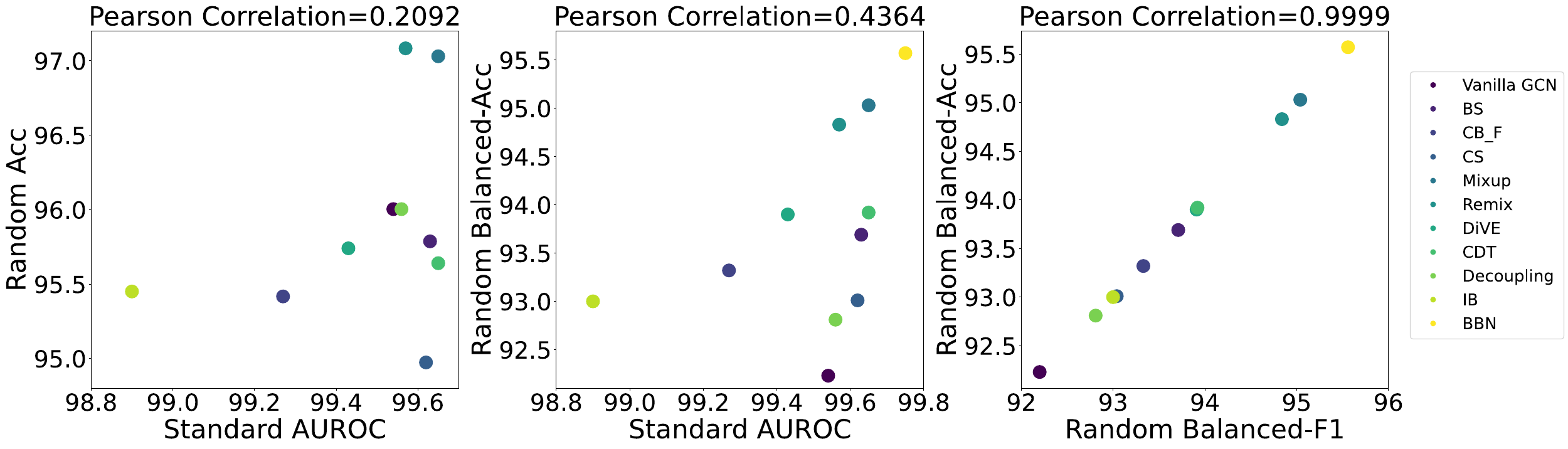}
    \caption{\textbf{Pearson's correlations} between the proposed metrics and the commonly used ones on USPTO-50K. Points in the figure are plotted for 11 baselines in Table~\ref{table:overall}.}
    \label{fig:corr}
\end{figure}

As measures of model discriminative power, accuracy and AUROC are the two most commonly used metrics for binary classification~\cite{huang2021therapeutics,2022arXiv220109637J,LT_Survey}. However, as shown in Table~\ref{table:overall}, they can be over-optimistic and insensitive to different models on highly imbalanced datasets. To overcome this challenge, ImDrug proposes to use balanced accuracy and balanced F1 score instead. To back up this design choice, we use reported AUROC on the \emph{balanced test set} of standard split as the gold standard or anchor points, and plot the corresponding values of accuracy as well as balanced accuracy on \emph{the imbalanced test set} of random split for comparison. Illustrated in Fig.~\ref{fig:corr}, AUROC on standard split achieves good agreement with \emph{balanced accuracy} on random split (Pearson's R $=0.4364$), whereas the correlation between that and \emph{regular accuracy} is significantly lower (Pearson's R $=0.2092$). Moreover, the two proposed balanced metrics on random split reach an excellent agreement (Pearson's R $=0.9999$), demonstrating the effectiveness of our novel balanced F1 score on imbalanced distributions as well.

%% file: Tables-new/table-1-new.tex
% \usepackage{multirow}

\begin{sidewaystable*}[thp]
\small
\centering
\caption{\textbf{Results for random and standard splits on 4 ImDrug classification datasets.} We perform binary classification on HIV~ (single-instance prediction) and SBAP~ (multi-instance prediction), and long-tailed classification on USPTO-50K~ (single-instance prediction) and DrugBank~ (multi-instance prediction). For each split and metric, the best method is \textbf{bolded} and the second best is \underline{underlined}.~ Four classical machine learning baselines with ECFP4~\cite{DBLP:journals/jcisd/RogersH10} input features are added, all of which under-perform the neural network-based methods. Some classes in DrugBank only have 1 sample, making over-sampling strategies like SMOTE~\cite{chawla2002smote} not applicable. Hence the results are missing.} \label{table:overall}

\begin{adjustbox}{max width=\textwidth}
\begin{tabular}{ccc|ccc|ccc|ccc|ccc}
\toprule
                                        \multicolumn{3}{c}{}  & \multicolumn{3}{c}{\textbf{HIV}\cite{wu2018moleculenet}}                                                                        & \multicolumn{3}{c}{\textbf{SBAP}\cite{2022arXiv220109637J}}         & \multicolumn{3}{c}{\textbf{USPTO-50K}\cite{liu2017retrosynthetic}}    & \multicolumn{3}{c}{\textbf{DrugBank}\cite{wishart2018drugbank}}     \\\midrule
\multicolumn{3}{c}{\# Graphs}                        & \multicolumn{3}{c}{41,127}                                                                     & \multicolumn{3}{c}{32,140}             & \multicolumn{3}{c}{50,036}             & \multicolumn{3}{c}{191,808}             \\
\multicolumn{3}{c}{\# Average nodes}                & \multicolumn{3}{c}{25.51}                                                                      & \multicolumn{3}{c}{32.60}             & \multicolumn{3}{c}{25.03}             & \multicolumn{3}{c}{26.43}             \\
\multicolumn{3}{c}{\# Average edges}         & \multicolumn{3}{c}{54.94}                                                                      & \multicolumn{3}{c}{103.94}             & \multicolumn{3}{c}{78.29}             & \multicolumn{3}{c}{83.21}             \\
\multicolumn{3}{c}{\# Classes}            & \multicolumn{3}{c}{2}                                                                          & \multicolumn{3}{c}{2}             & \multicolumn{3}{c}{10}             & \multicolumn{3}{c}{86}  \\           
\multicolumn{3}{c}{Imbalance ratio}            & \multicolumn{3}{c}{27.50}                                                                          & \multicolumn{3}{c}{36.77}             & \multicolumn{3}{c}{65.78}             & \multicolumn{3}{c}{216.96} 
\\ \midrule\midrule
\multicolumn{3}{c|}{Metric}                                                                                         & Balanced-Acc                          & Balanced-F1                           & AUROC            & Balanced-Acc & Balanced-F1 & AUROC & Balanced-Acc & Balanced-F1 & AUROC & Balanced-Acc & Balanced-F1 & AUROC \\ \midrule
\multicolumn{1}{c|}{}                            &                                            & Vanilla GCN       & 71.89                        & 69.82                         & 83.03 &  77.50            &     76.39        &  95.07   & 92.23      & 92.20            & 99.76     & 87.54     & 87.60     & 99.78      \\\cmidrule{2-15}
\multicolumn{1}{c|}{}                            &                       &  Random Forest       &  63.84                        &  58.52    &  82.84 &   63.33           &      57.98        &   92.49   &  70.10      &  70.43            &  98.33     &  60.94     &  63.65     &  98.77      \\
\multicolumn{1}{c|}{}                            &                                            &  Desicion Tree       &   57.98    &  49.13 &  64.52 &   51.72            &     37.12        &   73.60   &  65.98      &  66.49            &  81.40     &  56.25     &  59.41     &  78.14      \\
\multicolumn{1}{c|}{}                            &                                            &  SMOTE~\cite{chawla2002smote}       &  66.68                        &  62.67  &  83.67 &   76.68            &      76.88        &   82.04   &  77.51      &  78.06            &  98.28     &  -    &  -     &  -      \\
\multicolumn{1}{c|}{}                            &               \multirow{-4}{*}{\begin{tabular}{c}ML-based\\(non-deep)\end{tabular}} &  k-NN       &  62.95  &  57.20  &  72.43 &   74.94            &      73.59        &   75.11   &  64.26      &  65.02            &  92.63     &  48.51     &  51.43     &  88.87      \\

\cmidrule{2-15}
\multicolumn{1}{c|}{}                            &                                            & BS  & \textbf{77.13}                        & \textbf{76.71}                        & \textbf{83.91} &   83.21          &      83.06       &  91.01   & 93.69     & 93.71    & 99.76      & \textbf{94.98}    & \textbf{94.77}          &  99.60   \\
\multicolumn{1}{c|}{}                            &                                            & CB\_F  & 73.14                        & 72.66                        & 82.23 &    81.02         &      79.88       &  94.59   & 93.32     & 93.33    & 99.49      & 93.71    & 93.79          &  99.31   \\
\multicolumn{1}{c|}{}                            & \multirow{-3}{*}{\begin{tabular}{c}Class- \\ rebalancing\end{tabular}}        & CS    & \underline{77.04}                        & \underline{76.66}                        & 79.90 &        \textbf{91.49}      &      \textbf{91.48}       & \textbf{96.67}   & 93.01     & 93.04    & 99.71      & 93.67    & 93.70       & 99.39     \\\cmidrule{2-15}
\multicolumn{1}{c|}{}                            &                                            & Mixup              & 73.06                       & 70.62                        & 82.99 &           78.38   &      77.41       &  95.81   & \underline{95.03}     & \underline{95.04}    & \underline{99.77}      & 92.15            & 91.78             &  \textbf{99.84}    \\
\multicolumn{1}{c|}{}                            &                                            & Remix              & 74.43                        & 72.19                        & 81.09 &    79.58          &      78.49       &  95.01   & 94.83     & 94.84    & 95.26     & 94.47              & \underline{94.32}             & 99.76     \\
\multicolumn{1}{c|}{}                            & \multirow{-3}{*}{\begin{tabular}{c}Information \\augmentation\end{tabular}} & DiVE               & 75.07 & 74.24 & 81.12 &         84.58     &        84.65     &   95.43    & 93.90      &  93.91           & 99.60     & 90.26 & 91.08 & 99.42  \\\cmidrule{2-15}
\multicolumn{1}{c|}{}                            &                                            & CDT                & 71.67                        & 69.52                         & 82.95 &        78.92      &     78.17        &  95.58   & 93.92     & 93.92   & \textbf{99.78} & \underline{94.54}              & \underline{94.32}             & 99.52     \\
\multicolumn{1}{c|}{}                            &                                            & Decoupling         & 74.64 &  73.33 & \underline{83.66} &      85.04        &       84.82      &    95.66 &  92.81        & 92.81           & 94.43     &  90.00   &  90.27           & \underline{99.79}     \\
\multicolumn{1}{c|}{}                            &                                            & IB & 73.80                        &         71.32                & 79.88 &         83.70     &      83.61       &  90.65   & 93.00      & 93.00   & 99.66     & 90.38              & 90.10             & 99.39     \\
\multicolumn{1}{c|}{\multirow{-16}{*}{Random}}   & \multirow{-4}{*}{\begin{tabular}{c}Module \\Improvement\end{tabular}}       & BBN                & 76.50                        & 76.36                        & 81.65 &       \underline{90.67}       &      \underline{90.67}       &   \underline{96.16}  & \textbf{95.57}      & \textbf{95.56}   & 99.67   &  92.55             & 92.54             & 99.48     \\ \midrule\midrule
\multicolumn{1}{c|}{}                            &                                            & Vanilla GCN       & 71.24                        & 68.99                        & 80.99 &    77.29          &  76.21           &  94.70   &  92.15      & 92.15  & 99.54   & 89.38              &    86.96         & 98.09     \\\cmidrule{2-15}
\multicolumn{1}{c|}{}                            &    
&  Random Forest       &  64.58                        &  59.50                         &  82.35 &   59.19            &      51.54        &   91.64   &  71.63      &  71.75            &  98.01     &  61.90     &  56.75     &  97.34      \\
\multicolumn{1}{c|}{}                            &                                            &  Desicion Tree       &  57.20                        &  48.01                         &  67.36 &   50.27            &      33.93        &   80.08   &  72.33      &  72.16            &  84.63     &  52.38     &  44.96     &  76.37      \\
\multicolumn{1}{c|}{}                            &                                            &  SMOTE~\cite{chawla2002smote}       &  67.23                        &  63.40                         &  82.91 &   70.27            &      68.65        &   92.62   &  79.07      &  79.56            &  98.33     &   -   &     -  &   -     \\
\multicolumn{1}{c|}{}                            &                \multirow{-4}{*}{ \begin{tabular}{c}ML-based\\(non-deep)\end{tabular}}                             &  k-NN       &  64.02                        &  58.79                         &  78.78 &   62.43            &      56.66        &   78.99   &  67.91      &  68.56            &  92.79     &  42.86     &  38.02     &  88.66      \\
\cmidrule{2-15}
\multicolumn{1}{c|}{}                            &                                            & BS  & 75.04                        & 74.85                        & 79.60 &     88.41         &      88.38       &  \underline{95.16}   & 93.62    & 93.59  & 99.63    & \underline{94.69}             & 92.58            & \underline{98.93}     \\
\multicolumn{1}{c|}{}                            &                                            & CB\_F  & 71.57                       & 69.57                    & 80.59 &         84.70     &       83.78      &  94.93     & 94.07  & 94.11  & 99.27  &   90.55           & 88.09             & 97.77     \\
\multicolumn{1}{c|}{}                            & \multirow{-3}{*}{\begin{tabular}{c}Class- \\rebalancing\end{tabular}}        & CS    & \textbf{76.95}                        & \textbf{76.39}                        & 81.06 &  \textbf{89.79}            &      \textbf{89.78}      & \textbf{95.61}   &  92.73  & 92.76 & 99.62 &   \underline{94.69}         & \underline{92.89}             & 98.50     \\\cmidrule{2-15}
\multicolumn{1}{c|}{}                            &                                            & Mixup              & 71.48                        & 69.31                        &  \underline{81.18} &      79.69       &      78.04       &   94.99  &  95.21      & 95.19 & \underline{99.65} &  85.45            & 83.48             & \textbf{99.13}     \\
\multicolumn{1}{c|}{}                            &                                            & Remix              & 71.59                        & 69.47                        & \textbf{81.53} &      77.90        &      76.71       & 94.56    & \underline{95.27}       & \underline{95.25} & 99.57     & 89.93              & 87.46             & 98.78     \\
\multicolumn{1}{c|}{}                            & \multirow{-3}{*}{\begin{tabular}{c}Information \\augmentation\end{tabular}} & DiVE               &  74.16 &  72.69 & 81.06 &   86.69           &      86.65       & 93.41    &    94.03      & 94.00     & 99.43     & 89.95  & 87.46 & 98.33 \\\cmidrule{2-15}
\multicolumn{1}{c|}{}                            &                                            & CDT                & 71.87                       & 68.94                        & 80.47 &       81.51       &       80.98      &    94.14  & 93.79      & 93.78 & \underline{99.65}   &  91.15         & 88.29             & 98.80     \\
\multicolumn{1}{c|}{}                            &                                            & Decoupling         & 73.14                        & 71.52                        & 80.00 &       69.58       &      66.33       &   90.99  &  92.64      & 92.60 & 99.56    & 89.97        & 87.30       & 98.61    \\
\multicolumn{1}{c|}{}                            &                                            & IB &  \underline{75.34} &  \underline{75.14} & 80.16 &     84.87         &       84.87      &    91.41  & 91.51   & 91.55  & 98.90     &    89.98          & 87.14             &  98.39    \\
\multicolumn{1}{c|}{\multirow{-16}{*}{Standard}} & \multirow{-4}{*}{\begin{tabular}{c}Module \\Improvement\end{tabular}}       & BBN                & 73.50                        & 73.39                        & 80.96 &        \underline{88.49}      &      \underline{88.45}       &  94.02   & \textbf{96.23}       & \textbf{96.25}  & \textbf{99.75}   & \textbf{95.28}             & \textbf{93.61}             & 98.52    \\ \bottomrule
\end{tabular}
\end{adjustbox}

\end{sidewaystable*}

%% file: Tables-new/table-3-new.tex
\begin{table*}[htbp]
  \centering
  \caption{\textbf{Ablations of backbone models on USPTO-50K and USPTO-Catalyst.} USPTO-Catalyst here was formulated as multi-instance prediction.}\label{table:backbones}
  \scalebox{1}{
    \begin{tabular}{c|cc|ccc|ccc}
    \toprule
    \multicolumn{1}{c}{} &       &       & \multicolumn{3}{c|}{USPTO-50k} & \multicolumn{3}{c}{USPTO-Catalyst} \\
    \multicolumn{1}{c}{} &       &       & Balanced-Acc & Balanced-F1 & AUROC & Balanced-Acc & Balanced-F1 & AUROC \\
    \midrule
    \midrule
    \multirow{12}[8]{*}{Random} & \multirow{3}[2]{*}{Baseline} & Morgan &$81.42_{\pm1.19}$ & $81.56_{\pm1.16}$ & $97.40_{\pm0.74}$ & $14.62_{\pm0.49}$ & $16.57_{\pm0.58}$ & $87.65_{\pm0.21}$  \\
          &       & Transformer &$91.54_{\pm0.47}$ & $91.62_{\pm0.46}$ & $99.52_{\pm0.09}$ &$17.11_{\pm1.34}$ & $17.08_{\pm3.84}$ & $\boldsymbol{94.72_{\pm0.16}}$  \\
          &       & GCN   & $92.23_{\pm0.87}$ & $92.20_{\pm0.90}$ & $\underline{99.76_{\pm0.06}}$ & $16.33_{\pm0.56}$ & $15.85_{\pm2.15}$ & $92.81_{\pm0.41}$  \\
\cmidrule{2-9}          & \multirow{3}[2]{*}{CB\_F} & Morgan & $81.59_{\pm0.44}$ & $81.85_{\pm0.44}$ & $95.61_{\pm0.37}$ & $13.14_{\pm1.44}$ & $14.67_{\pm1.52}$ & $85.82_{\pm0.27}$  \\
          &      & Transformer &$91.49_{\pm0.66}$ & $91.55_{\pm0.62}$& $99.20_{\pm0.13}$ & $17.98_{\pm0.38}$ & $18.17_{\pm0.45}$ & $90.45_{\pm0.08}$  \\
          &       & GCN   & $93.32_{\pm0.18}$ & $93.33_{\pm0.20}$ & $99.49_{\pm0.10}$ & $\underline{21.36_{\pm0.61}}$ & $\boldsymbol{23.09_{\pm0.72}}$ & $90.56_{\pm0.10}$  \\
\cmidrule{2-9}          & \multirow{3}[2]{*}{Mixup} & Morgan & $83.72_{\pm0.56}$ & $83.88_{\pm0.53}$ & $97.66_{\pm0.23}$ & $15.41_{\pm0.3}$ & $16.39_{\pm0.36}$ & $89.38_{\pm0.17}$  \\
          &      & Transformer & $91.22_{\pm0.61}$ & $91.29_{\pm0.57}$ & $99.16_{\pm0.13}$ & $12.86_{\pm0.18}$ & $12.57_{\pm0.19}$ & $\underline{94.34_{\pm0.06}}$  \\
          &       & GCN   &$\underline{95.03_{\pm0.77}}$ & $\underline{95.04_{\pm0.78}}$ & $\boldsymbol{99.77_{\pm0.03}}$ & $16.95_{\pm0.56}$ & $18.69_{\pm0.69}$& $94.11_{\pm0.09}$ \\
\cmidrule{2-9}          & \multirow{3}[2]{*}{BBN} & Morgan & $83.57_{\pm1.32}$ & $83.81_{\pm1.43}$ & $97.11_{\pm0.14}$ & $18.30_{\pm0.65}$ & $21.18_{\pm0.66}$ & $83.59_{\pm0.39}$  \\
          &       & Transformer & $89.51_{\pm0.81}$ & $89.61_{\pm0.77}$ & $98.41_{\pm0.46}$ & $9.71_{\pm8.32}$ & $9.03_{\pm7.83}$ & $76.53_{\pm22.26}$  \\
          &       & GCN   &  $\boldsymbol{95.57_{\pm0.14}}$ & $\boldsymbol{95.56_{\pm0.14}}$ & $99.67_{\pm0.03}$ & $\boldsymbol{22.27_{\pm1.73}}$ & $\underline{22.70_{\pm1.40}}$ & $89.08_{\pm1.86}$  \\
    \midrule
    \midrule
    \multirow{12}[8]{*}{Standard} & \multirow{3}[2]{*}{Baseline} & Morgan & $82.04_{\pm3.13}$ & $81.77_{\pm3.41}$ & $96.94_{\pm0.60}$ & $13.89_{\pm0.17}$ & $12.96_{\pm1.95}$ & $87.79_{\pm0.73}$  \\
          &       & Transformer &  $92.13_{\pm1.02}$ & $92.17_{\pm0.99}$ & $99.47_{\pm0.35}$  & $16.66_{\pm0.90}$ & $15.68_{\pm2.40}$ & $\boldsymbol{94.97_{\pm0.36}}$  \\
          &       & GCN   & $92.15_{\pm0.88}$ & $92.15_{\pm0.87}$ & $99.54_{\pm0.12}$& $16.33_{\pm0.56}$ & $15.85_{\pm2.15}$ & $92.81_{\pm0.41}$  \\
\cmidrule{2-9}          & \multirow{3}[2]{*}{CB\_F} & Morgan &  $81.09_{\pm4.14}$ & $81.16_{\pm4.18}$ & $94.20_{\pm0.36}$ & $12.51_{\pm0.56}$ & $11.76_{\pm1.43}$ & $86.48_{\pm0.47}$  \\
          &       & Transformer &  $91.78_{\pm1.13}$ & $91.86_{\pm1.13}$ & $99.21_{\pm0.23}$  & $18.26_{\pm0.72}$ & $16.48_{\pm1.50}$ & $91.19_{\pm0.51}$  \\
          &       & GCN   & $94.07_{\pm1.18}$ & $94.11_{\pm1.20}$ & $99.27_{\pm0.31}$ & $\underline{21.68_{\pm0.28}}$ & $\underline{20.51_{\pm1.93}}$ & $90.82_{\pm0.40}$  \\
\cmidrule{2-9}          & \multirow{3}[2]{*}{Mixup} & Morgan & $84.26_{\pm1.26}$ & $84.33_{\pm1.35}$ & $96.78_{\pm0.97}$ & $14.86_{\pm0.10}$  & $13.45_{\pm1.37}$ & $89.50_{\pm0.49}$  \\
          &      & Transformer &$91.58_{\pm1.51}$ & $91.64_{\pm1.49}$ & $98.74_{\pm0.13}$  & $12.41_{\pm0.17}$  & $10.45_{\pm0.89}$ & $\underline{94.90_{\pm0.27}}$  \\
          &       & GCN   &$\underline{95.21_{\pm0.37}}$ & $\underline{95.19_{\pm0.33}}$ & $\underline{99.65_{\pm0.29}}$ & $16.62_{\pm1.03}$ & $15.50_{\pm2.28}$ & $94.38_{\pm0.28}$  \\
\cmidrule{2-9}          & \multirow{3}[2]{*}{BBN} & Morgan &  $85.26_{\pm2.78}$ & $85.31_{\pm2.89}$ & $97.55_{\pm0.61}$ & $17.04_{\pm1.69}$ & $16.5_{\pm3.40}$ & $86.30_{\pm0.37}$  \\
          &       & Transformer &$88.77_{\pm2.81}$ & $89.02_{\pm2.68}$ & $98.63_{\pm0.18}$& $14.25_{\pm2.14}$ & $12.19_{\pm2.91}$ & $91.36_{\pm1.12}$  \\
          &       & GCN   & $\boldsymbol{96.23_{\pm1.35}}$ & $\boldsymbol{96.25_{\pm1.32}}$ & $\boldsymbol{99.75_{\pm0.23}}$  & $\boldsymbol{21.79_{\pm0.96}}$ & $\boldsymbol{20.61_{\pm1.69}}$ & $91.18_{\pm1.70}$  \\
    \bottomrule
    \end{tabular}%
    }
\end{table*}%

%% file: Tables-new/table-2-new.tex
\begin{table*}[t]
\caption{\textbf{Results on Open LT under standard split.} The proportions of open classes in USPTO-50K (single-instance prediction) and DrugBank (multi-instance prediction) are both 0.2.}\label{table:openLT}
\centering
\small
\setlength{\tabcolsep}{4pt}
\scalebox{1}{
\begin{tabular}{c|ccc|cll}
\toprule
                  & \multicolumn{3}{c|}{\textbf{USPTO-50K}}           & \multicolumn{3}{c}{\textbf{DrugBank}}                                                 \\ \cmidrule{2-7} 
                  & Balanced-Acc & Balanced-F1 & AUROC     & Balanced-Acc & \multicolumn{1}{c}{Balanced-F1} & \multicolumn{1}{c}{AUROC} \\ \midrule
Vanilla GCN      & $85.82_{\pm0.42}$   & $81.38_{\pm0.41}$  & $94.18_{\pm0.066}$ & $91.72_{\pm0.28}$   & $91.75_{\pm0.33}$                      & $99.07_{\pm0.015}$                  \\\midrule
BS & $85.88_{\pm1.02}$   & $81.44_{\pm0.90}$  & $94.22_{\pm0.68}$  & $94.71_{\pm0.044}$   & $94.14_{\pm0.05}$                      & $99.11_{\pm0.03}$                  \\
CB\_F & $86.77_{\pm0.99}$   & $82.20_{\pm1.22}$  & $94.16_{\pm0.08}$  & $93.17_{\pm0.43}$   & $92.84_{\pm0.40}$                      & $98.80_{\pm0.11}$                  \\
Remix             & $87.92_{\pm0.23}$   & $83.40_{\pm0.21}$  & $93.91_{\pm0.13}$  & $92.68_{\pm0.83}$   & $92.75_{\pm0.59}$                     & $99.59_{\pm0.40}$                  \\
IB & $86.06_{\pm1.14}$   & $81.63_{\pm1.09}$   & $94.15_{\pm0.040}$  & $92.95_{\pm0.31}$   & $92.31_{\pm0.31}$                     & $99.03_{\pm0.04}$                  \\
BBN               & $87.86_{\pm0.14}$   & $83.32_{\pm0.13}$  & $94.07_{\pm0.19}$  & $94.52_{\pm0.33}$   & $93.96_{\pm0.34}$                      & $98.92_{\pm0.07}$                  \\ \midrule
OLTR         & ${89.21_{\pm0.75}}$   & $85.11_{\pm0.38}$  & ${96.69_{\pm0.12}}$  & $95.37_{\pm0.30}$   & $94.92_{\pm0.16}$                      & $\boldsymbol{99.79_{\pm0.09}}$                  \\
IEM          & $89.99_{\pm0.29}$   & $85.46_{\pm0.20}$  & $\underline{96.74_{\pm0.09}}$  & $95.77_{\pm0.16}$   & $95.16_{\pm0.15}$                      & $99.64_{\pm0.08}$                  \\ 
OLTR+BBN   &            $\boldsymbol{91.18_{\pm0.11}}$     &  $\boldsymbol{87.97_{\pm0.60}}$                  &    $\boldsymbol{97.72_{\pm0.56}}$                            &              $\boldsymbol{96.84_{\pm1.59}}$                   &               $\boldsymbol{96.12_{\pm0.93}}$                 &                     $\boldsymbol{99.89_{\pm0.69}}$                                                                                       \\
IEM+BBN    &          $\underline{90.66_{\pm0.41}}$                           &         $\underline{86.03_{\pm0.75}}$                       &           $96.64_{\pm1.03}$                      &            $\underline{96.06_{\pm0.76}}$                    &                  $\underline{95.59_{\pm0.96}}$                                  &    $99.77_{\pm0.12}$                                                    \\
\bottomrule
\end{tabular}
}
\end{table*}

%% file: Sections-TPAMI/5-Conclusion.tex
\section{Conclusion}
\label{sec:conclusion}
In this work, motivated by the observed prevalence of imbalanced distribution of the real-world pharmaceutical datasets, we present ImDrug, a comprehensive platform for systematic model development and benchmarking to facilitate future research in AIDD. Unfortunately, targeted at the intersection of AIDD and deep imbalanced learning, we find that this subject has not received sufficient attention by the vast majority of work on ML modeling in either field alone. Through extensive experiments, it is shown that the existing algorithms as well as evaluation metrics leave sufficient room for improvement to be truly applicable in practice. We hope that this paper offers valuable insight for the machine learning and medicinal chemistry community as a whole, to think more critically about the importance and challenges of building trustworthy solutions to real-world AIDD problems. Continued development of ImDrug such as incorporating 3D-structured features, pre-trained model baselines, novel metrics for imbalanced regression and benchmarks for generation tasks will be supported to enrich this interdisciplinary and promising area of research.

%% file: Appendix/Appendix-A-Accessibility.tex
\section{Key Information about ImDrug}
\subsection{Dataset Documentation}
For each dataset, we provide the corresponding description which includes data statistics, data source, unit, and references. 
Please refer to Appendix~\ref{appendix:datasets} for more details for all 11 datasets.
We host the ImDrug datasets at \href{https://doi.org/10.7910/DVN/TEXUKR}{Harvard Dataverse} and \href{https://drive.google.com/drive/folders/16dSuqq-Fh6iGqjPL1phtQT3C_K70cCfK?usp=sharing}{Google Drive}, and the data is instantly accessible in human-readable form without featurization. 

\subsection{Intended Uses}
ImDrug is intended for researchers in biomedical, machine learning, and data science to facilitate interdisciplinary research for AI-aided Drug Discovery (AIDD) and deep imbalanced learning.

\subsection{Hosting and Maintenance Plan}
The ImDrug codebase is hosted and version-tracked via GitHub and it will be permanently available under the link \url{https://github.com/DrugLT/ImDrug}. All of the datasets are hosted at \href{https://doi.org/10.7910/DVN/TEXUKR}{Harvard Dataverse} and \href{https://drive.google.com/drive/folders/16dSuqq-Fh6iGqjPL1phtQT3C_K70cCfK?usp=sharing}{Google Drive} for public access and download. 

ImDrug is a community-driven and open-source initiative. Our core development team will be committed to the maintenance and development of the benchmarks and datasets in the next five years at minimum. We plan to grow ImDrug by introducing new learning tasks, datasets, novel baselines, competitive backbones, and leaderboards. We welcome external contributors.

\subsection{Licensing}
% We license ImDrug's Python package using the MIT license.
ImDrug codebase is under the MIT License. For individual dataset usage, please refer to the dataset license found in the website.

\subsection{Author Statement}
We the authors bear all responsibility in case of violation of rights. 

\subsection{Computing Resources}
We use a computing server with NVIDIA Tesla V100 GPUs (32GB) and Inter(R) Xeon(R) CPUs for all empirical experiments in this paper. Each trial of experiment is run on 1 V100 GPU and 8 CPU cores in the docker image built by the docker file in \url{https://github.com/DrugLT/ImDrug}. For more detailed information, please refer to Appendix~\ref{appendix:G.3}.

\subsection{Limitations}
ImDrug includes commonly-used methods and datasets targeted at the intersection of imbalanced learning and AIDD, which make up the proposed emerging cross-discipline. However, ImDrug is an ongoing effort and we strive to continuously include more novel baselines that are not only explored in the field of Computer Vision. In addition, more new datasets and tasks for AIDD will also be included as follow-up work, such as the molecule generation tasks involving de novo drug design and retrosynthesis. 

\subsection{Potential Negative Societal Impacts}
AIDD is an emerging area of research with high potential to revolutionize the pharmaceutical industry by expediting the development of safe and effective drugs. 
Our proposed ImDrug benchmark does not involve human subjects research or any personally identifiable information. However, for adaptation of some of the imbalanced learning baselines originally proposed in Computer Vision to the AIDD domain, we choose certain drug/protein encodings and model backbones which may not realize their full potential due to the lack of best practices. Moreover, even though we tried to make the datasets as comprehensive as possible, their current form in limited scale and scope are likely to fall short in representing some of the real-world distributions and scenarios. Being aware of these limitations, we encourage the users and followers of this work to explore better implementation protocols, and in the mean time the datasets and trained models should be used with caution especially when applied to real-world problems.
% Direct usage of machine learning models in a clinic environment without prior validations will easily lead to negative outcomes. 

%% file: Appendix/Appendix-B-Proof.tex
\section{Theoretical Proofs}\label{sec:appendix:proof}

\begin{theorem}\label{thm:appendix:balanced-f1}
Let $Rec_k$ and $Prec_k$ denote the recall and balanced precision for class $k$. Given a trained predictor, the evaluated balanced F1 score

\begin{equation}
    \text{Balanced-F1} \coloneqq \frac{1}{K}\sum_{k=1}^{K}\frac{2\times Rec_k\times Prec_k}{Rec_k + Prec_k} \label{eqn:appendix:balanced-f1}
\end{equation}

is invariant on any test set whose samples of each class are drawn i.i.d from a fixed distribution. 
\end{theorem}

\begin{proof}
% Given a trained predictor $P$, we denote by $P(k|j)$ the probability of classifying a sample of ground truth label $j$ to class $k$, $i.e.$,

% \begin{equation}
%     P(j|k) = \mathbb{E}\left[\frac{\sum_{i=1}^n\mathbbm{1}(y_i=j,\hat{y}_i=k)}{\sum_{i=1}^n\mathbbm{1}(\hat{y}_i=k)}\right]
We start by formalizing Definition 3.1
% Def.~\ref{def:balanced_metrics}
from a probabilistic view. Consider the test set $\{x_i, y_i\}_{i=1}^n$ as $n$ $i.i.d$ draws of random variables $x\in\mathcal{X}$ and $y\in\mathcal{Y}$ from the distribution $P(x)$ and $P(y|x)$ respectively. And the trained classifier $g:\mathcal{X}\rightarrow\mathcal{Y}$ makes the corresponding set of predictions $\{\widehat{y}_i=g(x_i)\}_{i=1}^n$. Then the conventional one-vs-all recall $C\text{-}Rec_k$ and precision $C\text{-}Prec_k$ for class $k$ follow

\begin{align}
    C\text{-}Rec_k &= \frac{\mathbb{E}\left[\sum_{i=1}^n\mathbbm{1}(y_i=k, \widehat{y}_i=k)\right]}{\mathbb{E}\left[\sum_{i=1}^n\mathbbm{1}(y_i=k)\right]} =  \frac{P(\widehat{y}=k, y=k)}{P(y=k)} = \frac{P(\widehat{y}=k|y=k)P(y=k)}{P(y=k)}\label{eqn:appendix:recall}\\
    C\text{-}Prec_k &= \frac{\mathbb{E}\left[\sum_{i=1}^n\mathbbm{1}(y_i=k, \widehat{y}_i=k)\right]}{\mathbb{E}[\sum_{i=1}^n\mathbbm{1}(\widehat{y}_i=k)]} =  \frac{P(\widehat{y}=k, y=k)}{P(\widehat{y}=k)} = \frac{P(\widehat{y}=k|y=k)P(y=k)}{P(\widehat{y}=k)}\label{eqn:appendix:precision}
\end{align}

% where we use the Bayes' Theorem for obtaining the last equality of Eqn.~\ref{eqn:recall} and \ref{eqn:precision}. 
Note that $C\text{-}Rec_k \equiv Rec_k$. Moreover, the balanced precision for class $k$, as in Eqn~\ref{eqn:balanced-prec}, can be written as 

\begin{align}
    Prec_k &=  \frac{\mathbb{E}\left[\sum_{i=1}^n\mathbbm{1}(y_i=k, \widehat{y}_i=k)\right]}{\mathbb{E}[\sum_{i=1}^n\mathbbm{1}(y_i=k, \widehat{y}_i=k) + \sum_{j\ne k}\sum_{i=1}^n\pi_{jk}\mathbbm{1}(y_i=j, \widehat{y}_i=k)]}\\
    & =  \frac{P(\widehat{y}=k|y=k)P(y=k)}{\int_{y'} \pi_{y'k}P(\widehat{y}=k, y')dy'}\\
    & = \frac{P(\widehat{y}=k|y=k)P(y=k)}{\int_{y'} P(\widehat{y}=k|y=y')P(y=y')\times\frac{P(y=k)}{P(y=y')}dy'}\label{eqn:appendix:marginal_frac1}\\
    & = \frac{P(\widehat{y}=k|y=k)}{\int_{y'}P(\widehat{y}=k|y=y')dy'}\label{eqn:appendix:marginal_frac2}
\end{align}

where to derive Eqn.~\ref{eqn:appendix:marginal_frac1}, we use the fact that $\forall j, k\in\mathcal{Y}$, $\pi_{jk} = n_k/n_j = P(y=k)/P(y=j)$. 
We now proceed by proving the following lemma:

\begin{lemma}\label{lemma:appendix:invariant_probability}
The conditional probabilities $P(\widehat{y}=k|y=k)$ and $P(\widehat{y}=k|y=y')$ are invariant/constant regardless of the dataset distribution $P(x)$ and $P(y)$, assuming the samples of each class are drawn $i.i.d$ from a fixed distribution, $P(x|y)$.
\end{lemma}

\begin{proof}
We write $P(\widehat{y}=k|y=k)$ and $P(\widehat{y}=k|y=y')$ as the marginal likelihood functions of $x$:

\begin{align}
    P(\widehat{y}=k|y=k) &= \int_xP(\widehat{y}=k|x)P(x|y=k)dx\label{eqn:appendix:marginal1}\\
    P(\widehat{y}=k|y=y') &= \int_xP(\widehat{y}=k|x)P(x|y=y')dx\label{eqn:appendix:marginal2}
\end{align}    

Note that given $x$ and the trained classifier $g$, the probability $P(\widehat{y}=k|x)=P(g(x)=k|x)\propto\delta(g(x), k)$ is a fixed Dirac delta function. Additionally, since the conditional distribution $P(x|y)$ is assumed to be fixed for all $y\in\mathcal{Y}$, the integrand of the RHS of Eqn.~\ref{eqn:appendix:marginal1} and \ref{eqn:appendix:marginal2} are constant regardless of the dataset distribution $P(x)$ and $P(y)$, so are the LHS. This completes the proof. 
\end{proof}

With Lemma.~\ref{lemma:appendix:invariant_probability}, evidently both the numerator and denominator of the RHS of Eqn.~\ref{eqn:appendix:marginal_frac2} are invariant regardless of the dataset distribution $P(x)$ and $P(y)$. This proves that the conventional recall in Eqn.~\ref{eqn:appendix:recall}, the balanced precision in Eqn.~\ref{eqn:appendix:marginal_frac2} and the proposed balanced F1 score in Eqn.~\ref{eqn:appendix:balanced-f1} are all invariant/constant under any dataset distribution. In other words, given a trained classifier, the proposed balanced metrics of ImDrug provide unbiased estimate of its multi-class classification efficacy on any imbalanced test set.

\end{proof}

%% file: Appendix/Appendix-C-Usage.tex
\section{Documentation and Usages of the ImDrug Benchmark}
The source code of ImDrug is available at \url{https://github.com/DrugLT/ImDrug}. The datasets in ImDrug are hosted at \url{https://drive.google.com/drive/folders/16dSuqq-Fh6iGqjPL1phtQT3C_K70cCfK?usp=sharing}. ImDrug is intended for comprehensive comparison among imbalanced learning methods. All necessary configurations for training and evaluation are stored as a JSON dictionary file following a fixed format. Listing~\ref{listing:json} provides a simple example, which integrates all of the steps including dataset curation, dataset loading, and algorithm configurations in just a few lines of code.

% \subsection{Use of Datasets}

\colorlet{punct}{red!60!black}
\definecolor{background}{HTML}{F1FCFD}
\definecolor{delim}{RGB}{20,105,176}
\colorlet{numb}{magenta!60!black}

\lstdefinelanguage{json}{
    basicstyle=\normalfont\ttfamily,
    numbers=left,
    numberstyle=\scriptsize,
    stepnumber=1,
    numbersep=8pt,
    showstringspaces=false,
    breaklines=true,
    frame=lrtb,
    backgroundcolor=\color{background},
    literate=
     *{0}{{{\color{numb}0}}}{1}
      {1}{{{\color{numb}1}}}{1}
      {2}{{{\color{numb}2}}}{1}
      {3}{{{\color{numb}3}}}{1}
      {4}{{{\color{numb}4}}}{1}
      {5}{{{\color{numb}5}}}{1}
      {6}{{{\color{numb}6}}}{1}
      {7}{{{\color{numb}7}}}{1}
      {8}{{{\color{numb}8}}}{1}
      {9}{{{\color{numb}9}}}{1}
      {:}{{{\color{punct}{:}}}}{1}
      {,}{{{\color{punct}{,}}}}{1}
      {\{}{{{\color{delim}{\{}}}}{1}
      {\}}{{{\color{delim}{\}}}}}{1}
      {[}{{{\color{delim}{[}}}}{1}
      {]}{{{\color{delim}{]}}}}{1},
}

\begin{lstlisting}[language=json, numbers=none, caption={Algorithm conﬁguration example.}, captionpos=b, label={listing:json}]
{
    "dataset": {
        "drug_encoding": "Transformer",
        "protein_encoding": "Transformer",
        "tier1_task": "single_pred",
        "tier2_task": "ADME",
        "dataset_name": "BBB_Martins",
        "split": {
            "method": "standard",
            "by_class": false
        }
    },
    "loss": {
        "type": "CrossEntropy"
    },
    "train": {
        "batch_size": 128,
        "combiner": {
            "type": "bbn_mix"
        },
        optimizer=dict(
            type='ADAM',
            lr=1e-3,
            momentum=0.9,
            wc=2e-4,
        ),
        two_stage=dict(
            drw=False,
            drs=False,
            start_epoch=10,
        )
    },
    "setting": {
        "type": "LT Classification",
        "num_class": 10
    },
    "use_gpu": true
}
\end{lstlisting}

% \subsection{Benchmarking Methods and Baseline Results}
Note that each configuration in the JSON file can be chosen as follows:
\begin{itemize}[leftmargin=20pt]
    \item \textcolor{blue}{"drug\_encoding"}: ["Morgan", "Pubchem', "Daylight", "rdkit\_2d\_normalized", "ESPF", "CNN", "CNN\_RNN", 
    "Transformer", "MPNN", "ErG", "DGL\_GCN", "DGL\_NeuralFP", "DGL\_AttentiveFP", "DGL\_GIN\_AttrMasking", "DGL\_GIN\_ContextPred"]
    \item \textcolor{blue}{"protein\_encoding"}: ["AAC", "PseudoAAC", "Conjoint\_triad", "Quasi-seq", "ESPF", "CNN", "CNN\_RNN", "Transformer"]
    \item \textcolor{blue}{"tier1\_task"}: ["single\_pred", "multi\_pred"], both are applicable for hybrid prediction.
    \item \textcolor{blue}{"tier2\_task"}: ["ADME", "TOX", "QM", "HTS", "Yields", "DTI", "DDI", "Catalyst", "ReactType"]
    \item \textcolor{blue}{"dataset\_name"}: ["BBB\_Martins", "Tox21", "HIV", "QM9", "USPTO-50K", "USPTO-Catalyst", "USPTO-1K-TPL", "USPTO-500-MT", "USPTO-Yields", "SBAP", "BindingDB\_Kd"]
    \item \textcolor{blue}{"split.method"}: ["standard", "random", "scaffold", "time", "combination", "group", "open-random", "open-scaffold", "open-time", "open-combination", "open-group"]
    \item \textcolor{blue}{"setting.type"}: ["Imbalanced Classification", "LT Classification", "Imbalanced Regression", "Open LT"]
\end{itemize}

We also store the filtered data file without featurization for easy access in a human-readable form~(\textit{e.g.}, EXECL file). 
In addition, guidance for training and evaluation on a new dataset is described in the README.md file of the Github repository. The entries of 'dataset' in the JSON dictionary specify the elements of the three-level hierarchical design for datasets described in Section 3.1.1.
% ~\ref{sec:3.1.1}.
% \textit{i.e.}, \emph{single-instance prediction}, \emph{multi-instance prediction}, and \emph{hybrid prediction}.

% By documenting the script in 'baseline' entry, a baseline for imbalanced learning is specified. Baseline results across most datasets is available in the baseline\_results dictionary \url{https://github.com/DrugLT/ImDrug/baseline_results}. We present the experimental results for all 11 baselines under the split methods of random or standard and 3 backbones of DGL\_GCN, Transformer and Morgan. 

%% file: Appendix/Appendix-D-Dataset.tex
\section{Datasets}\label{appendix:datasets}

In this section, we give a description of the datasets we used in ImDrug. 
As shown in Table 2,
% ~\ref{table:datasets}, 
ImDrug consists of 11 imbalanced drug datasets, 8 task types, and 6 data split functions.
In what follows, we briefly introduce these datasets followed by their task types.

\subsection{Datasets Descriptions}

\textbf{ImDrug.BBB\_Martins} is a binary classification dataset~\cite{wu2018moleculenet}.
Given a drug SMILES string, the task is to predict the activity of the blood-brain barrier~(BBB). 
It contains 1,975 entities, each consisting of the drug name, the drug SMILES string, and a binary label indicating whether this drug can penetrate the blood-brain barrier.
The recommended setting of this dataset is imbalanced classification.

\textbf{ImDrug.Tox21} is a binary classification dataset~\cite{huang2021therapeutics}.
Given a drug SMILES string, the task is to predict the toxicity in one of 12 quantitative high throughput screening~(qHTS) assays.
It contains 7,831 entities, each consisting of the drug ID, the drug SMILES string, and 12 binary labels called NR-AR, NR-AR-LBD, NR-AhR, NR-Aromatase, NR-ER, NR-ER-LBD, NR-PPAR-gamma, SR-ARE, SR-ATAD5, SR-HSE, SR-MMP, SR-p53.
The recommended setting of this dataset is imbalanced classification.

\textbf{ImDrug.HIV} is a binary classification dataset~\cite{wu2018moleculenet}. 
Given a drug SMILES string, the task is to predict this drug's activity against the HIV virus.
It contains 41,127 entities, each consisting of the drug ID, the drug SMILES string, and a binary label indicating the activity of this drug against HIV.

\textbf{ImDrug.QM9} is a regression dataset~\cite{qm9_1, qm9_2}.
Given a drug 3D Coulomb matrix, the task is to predict 12 drug properties.
It contains 133,885 entities, each consisting of the drug ID, the drug SMILES string, the 3D coordinates of each atom, and 12 regression labels called \emph{Mu}, \emph{Alpha}, \emph{Homo}, \emph{Lumo}, \emph{Gap}, \emph{R2}, \emph{Zpve}, \emph{Cv}, \emph{U0}, \emph{U298}, \emph{H298}, \emph{G298}.
The recommended setting of this dataset is imbalanced regression.

\textbf{ImDrug.SBAP} is a binary classification dataset.
Given the amino acid sequence and the drug SMILES string, the task is to predict binding affinity between them.
It contains 32,140 entities, each consisting of the drug ID, the drug SMILES string, the amino acid sequence, the protein ID, and a binary label representing binding affinity.
It was extracted from the BindingDB~\cite{huang2021therapeutics} by DrugOOD~\cite{2022arXiv220109637J} for the interaction prediction task.
The recommended setting of this dataset is imbalanced classification.

\textbf{ImDrug.USPTO-Catalyst} is a multi-class classification dataset.
Given the set of reactants and products, the task is to predict the catalyst type.
It contains 721,799 entities, each consisting of the reactant ID, the reactant SMILES string, the product ID, the product SMILES string, and the catalyst type.
It was derived from the USPTO database by TDC~\cite{huang2021therapeutics} for the catalyst prediction task.
The recommended settings of this dataset are LT classification and Open LT.

\textbf{ImDrug.USPTO-1K-TPL} is a multi-class classification dataset.
Given a reaction SMILES string, the task is to predict the reaction type.
It contains 445,115 entities, each consisting of the reaction SMILES string, and the reaction type.
It was collated from the USPTO dataset by this work~\cite{schwaller2021mapping} for the reaction type classification task.
The recommended settings of this dataset are LT classification and Open LT.

\textbf{ImDrug.USPTO-Yields} is a regression dataset.
Given the set of reactants and products, the task is to predict the yields.
It contains 853,638 entities, each consisting of the reaction SMILES string, and the reaction yields.
It was derived from the USPTO database by TDC~\cite{huang2021therapeutics} for the yield prediction task.
The recommended setting of this dataset is imbalanced regression.

\textbf{ImDrug.USPTO-500-MT} has multiple reaction prediction tasks, including reaction yield prediction task, catalyst prediction task, and reaction type classification task.
It contains 143,535 entities, each consisting of the reaction SMILES string, the reaction type, and the reaction yields.
It was extracted from the USPTO dataset by T5Chem~\cite{lu2022unified} for multiple tasks.
The recommended settings of this dataset are LT classification/Open LT or imbalanced regression.

\textbf{ImDrug.USPTO-50K} is a multi-class classification dataset~\cite{huang2021therapeutics}.
Given the reaction SMILES strings, the task is to predict the reaction type.
It contains 50,016 entities, each consisting of the reaction SMILES string, and the reaction type.
The recommended settings of this dataset are LT classification and Open LT.

\textbf{ImDrug.DrugBank} is a multi-class classification dataset.
Given the SMILES strings of two drugs, the task is to predict the interaction type between them.
It contains 191,808 entities, each consisting of two drugs' SMILES strings, ID, and interaction type.
It was collated from FDA and Health Canada drug labels as well as from the primary literature by TDC~\cite{huang2021therapeutics} for the interaction type prediction task.
The recommended settings of this dataset are LT classification and Open LT.

%% file: Appendix/Appendix-E-Baseline.tex
\section{Imbalanced Learning Baselines}
\label{sec:appendix:baselines}
In this section, we provide detailed descriptions of the imbalanced learning baselines benchmarked in ImDrug.

\subsection{Baselines of Imbalanced \& Long-Tailed Classification}\label{sec:appendix:classification_baselines}

\textbf{Class Re-balancing.} \quad 4 imbalanced learning baselines are included in this category. Compared to the conventional \textbf{softmax cross-entropy loss (CE)}, \textbf{cost-sensitive loss (CS)}~\cite{CostSensitiveCE} re-weights the log likelihood of each prediction inversely proportional to its label frequency $\pi_y$. \textbf{Class-balanced loss}~\cite{ClassBalanceFocal} introduces a novel concept, namely effective number, to approximate the expected sample number of each class. The re-weighting factor is given by the reciprocal of the effective number in the form of an exponential function of the training sample number, which can be applied to a normal Focal loss~\cite{lin2017focal} (\textbf{CB\_F}) or cross-entropy loss (\textbf{CB\_CE}). \textbf{Balanced softmax}~\cite{BalancedSoftmaxCE} \textbf{(BS)} proposes to adjust the predicted logits by label frequencies to alleviate the bias of class imbalance. \textbf{Influence-balanced training}~\cite{park2021influence} \textbf{(IB)} down-weights highly influencial samples measured by the gradient magnitude at fine-tuning to smooth the decision boundary, thus mitigating over-fitting and class bias.
 
\textbf{Information Augmentation.} \quad ImDrug benchmarks 2 data augmentation (Mixup and Remix) and 1 transfer learning (DiVE) baselines.  \cite{zhang2018mixup}
\textbf{Mixup}~\cite{zhang2018mixup} is a classical trick which constructs augmented data by making convex combination of two samples and their labels to improve the generalization ability of models.
% The newly produced data distribution of Mixup is expressed as:
% \begin{equation}
%     \mu(\widetilde{x},\widetilde{y}|x_i,y_i)=\frac{1}{n}\sum_j^n\mathop{\mathbb{E}}\limits_\lambda[\delta(\widetilde{x}=\lambda\cdot x_i+(1-\lambda)\cdot x_j,\widetilde{y}=\lambda\cdot y_i+(1-\lambda)\cdot y_i)],
% \end{equation}
% where $(x_i,y_i)$ and $\lambda$ is randomly sampled from the beta distribution. In a nutshell,
Specifically, Mixup is formed by the linear interpolation of two samples $(x_i, y_i)$ and $(x_j, y_j)$ obtained at the training data:
\begin{equation}
\begin{aligned}
    \widetilde{x}^{MU}&=\lambda x_i+(1-\lambda)x_j\\
    \widetilde{y}^{MU}&=\lambda y_i+(1-\lambda)y_j \;,
\end{aligned}
\end{equation}
where $\lambda$ is randomly sampled from the predefined beta distribution.
Built upon Mixup, \textbf{Remix}~\cite{chou2020remix} assigns the mixed label in favor of the minority class by providing a disproportionately higher weight to the minority class. The formulation of Remix is as below:
\begin{equation}
\begin{aligned}
    \widetilde{x}^{RM}&=\lambda_xx_i+(1-\lambda_x)x_j\\
    \widetilde{y}^{RM}&=\lambda_yy_i+(1-\lambda_y)y_j \;,
\end{aligned}
\end{equation}
where $\lambda_x$ is sampled from the beta distribution and $\lambda_y$ is designed related to number of samples. 
% By doing so, the classifier learns to push the decision boundaries towards the majority classes and balance the generalization error between majority and minority classes.

However, it is worth noticing that the conventional Mixup/Remix approach performed on the input data in the CV domain cannot be trivially applied to graph or sequence data~\cite{G-Mixup} due to their non-Euclidean structures. Instead, manifold Mixup/Remix~\cite{verma2019manifold} which perform interpolations of hidden representations, are implemented in ImDrug. 

Lastly, as a transfer learning method, \textbf{DiVE}~\cite{DiVE} employs a class-balanced model as the teacher to generate virtual examples. By distillation, it achieves remarkable head-to-tail knowledge transfer for long-tailed learning. Supposing there are $C$ classes in total, we define the predicted logits of the student network and teacher network in DiVE as $\boldsymbol{s}=(s_1,s_2,...,s_C)$ and $\boldsymbol{t}=(t_1,t_2,...,t_C)$ respectively. The loss function of the student network is shown below:
\begin{equation}
    L_{KD}=-(1-\Lambda)\sum_{k=1}^Cy_k\text{log}s_k+\Lambda \sum_{k=1}^Ct_k\text{log}\frac{t_k}{s_k} \;,
\end{equation}
where the hyperparameter $\Lambda\in[0,1]$ balances the two terms.

\textbf{Module Improvement.} \quad We consider 1 classifier design (CDT), 1 decoupled training (Decoupling) and 1 ensemble learning (BBN) methods. \textbf{CDT}~\cite{CDT} proposes to incorporate class-dependent temperatures to force minor classes to have larger
decision values in the training phase, so as to compensate for the effect of feature
deviation in the test data. 
We denote $\boldsymbol{w}_c^{\top}\boldsymbol{f}_\theta(x)$ as the decision values of a training instance and $a_c$ as the temperature factor. The proposed training objective in CDT is as follows:

\begin{equation}
    -\sum_n\text{log}\left(  \frac{\text{exp}\left(  \frac{\boldsymbol{w}^{\top}_{y_n}\boldsymbol{f}_\theta(x_n)}{a_{y_n}} \right)}{\sum_c\text{exp}\left( \frac{\boldsymbol{w}^{\top}_{c}\boldsymbol{f}_\theta(x_n)}{a_c}  \right)} \right),
\end{equation}
where $c\in\{1,...,C\}$, $n\in\{1,...,N\}$, $C$ and $N$ denote the number of class and the number of training samples, respectively.

\textbf{Decoupling}~\cite{kang2020decoupling} is the pioneering work to introduce a two-stage training scheme. It employs instance-balanced sampling for representation learning in the first stage and transitions to class-balanced sampling for training the classifier in the later stage. In a nutshell, the combined sampling method is as follows:
\begin{equation} 
    p_j^{\text{PB}}(t)=(1-\frac{t}{T})p^{\text{IB}}_j+\frac{t}{T}p^{\text{CB}}_j,
\end{equation}
where $t$ is the current number of training epoch, $T$ is the overall epoch number, $p^{\text{IB}}$ means the instance-balanced sampling and $p^{\text{CB}}$ means the class-balanced sampling.

\textbf{BBN}~\cite{zhou2020bbn} consists of a conventional learning branch and a re-balancing branch. To handle long-tailed recognition, the predictions of two branches are dynamically combined during training, ensuring that the learning focus gradually shifts from head classes to tail classes. By using uniform and reversed samplers in the bilateral branches, two samples $(x_c,y_c)$ and $(x_r,y_r)$ are obtained as the input data. The output logits are as follows:

\begin{equation}
    \boldsymbol{z}=\beta \boldsymbol{W}_c^{\top}\boldsymbol{f}_c+(1-\beta)\boldsymbol{W}_r^{\top}\boldsymbol{f}_r,
\end{equation}
where the weights $\boldsymbol{f}_c$ and $\boldsymbol{f}_r$ are controlled with a trade-off parameter $\beta$, $\boldsymbol{W}_c$ and $\boldsymbol{W}_r$ mean the classifiers of the two branches. Finally, a loss of weighted cross-entropy classification is applied as:

\begin{equation}
    \mathcal{L}=\beta E(\hat{\boldsymbol{p}},y_c)+(1-\beta)E(\hat{\boldsymbol{p}},y_r).
\end{equation}

\subsection{Baselines of Open LT \& Imbalanced Regression}

\textbf{Open LT} can be regarded as a variant of conventional long-tailed classification with unseen and out-of-distribution (OOD) tail classes in the test set. In principle, all aforementioned baselines for imbalanced classification can be seamlessly transferred to this setting. However, the extra challenge posed by the protocol requires targeted remedies to OOD generalization to achieve state-of-the-art performance. Hence ImDrug includes 2 additional baselines for Open LT. \textbf{OLTR}~\cite{liu2019large} as the seminal work in this line of research, explores the idea of feature prototypes to handle long-tailed recognition with open-set detection. For an input drug/protein, OLTR first learns the visual memory $M$ of all the training data:
\begin{equation}
    M=\{c_i\}_{i=1}^K ,
\end{equation}
where $K$ is the number of training classes and $c_i$ is the centroid of each class group. The most important part is to differentiate the samples of the training dataset from those of open-set. 
% Let $v^{direct}$ be the feature vector to be discriminated, 
OLTR minimizes the distance between the $v^{direct}$ feature vector and the discriminative centroids:
\begin{equation}
    \gamma:=\text{reachability}(v^{direct},M)=\mathop{\text{min}}\limits_i\|v^{direct}-c_i \|_2.
\end{equation}

\textbf{IEM}~\cite{zhu2020inflated} further innovates the meta-embedding memory by a dynamical update scheme, where each class has independent memory blocks and records only the most discriminative feature prototypes. The soft attention mechanism is applied in IEM. Given a query \textbf{q}, the output $p$ is generated by:
\begin{equation}
    p=\frac{\sum_is(\textbf{q},\textbf{k}_i)v_i}{\sum_is(\textbf{q},\textbf{k}_i)},
\end{equation}
where $v_i$ denotes the $i$-th prediction score in value memory, and $\textbf{k}_i$ is the $i$-th vector of the key memory. $s(\cdot)$ denotes the similarity function that measures distances between two vectors. Another key component is the self-attention module. Define the query, key, and value as \textbf{Q}, \textbf{K} and \textbf{V} respectively, the global representation for the whole feature map is as follows:
\begin{equation}
    \text{SA}(\textbf{Q},\textbf{K},\textbf{V})=Softmax(\frac{\textbf{Q}\textbf{K}^T}{\sqrt{d}})\textbf{V},
\end{equation}
where $d$ is the size of input channel.

\textbf{Imbalanced Regression} can be reduced to conventional imbalanced classification by naively divide the continuous label space into multiple consecutive bins as classes. In this way, many baselines in Appendix~\ref{sec:appendix:classification_baselines} can be adapted by replacing the classification head as a regression head. However, such trivial transformation undermines the intrinsic topology of labels induced by the Euclidean distance, leading to sub-optimal performance~\cite{yang2021delving}. In ImDrug, we experiment with 3 extra baselines specialized at solving this particular challenge. \textbf{Focal-R} is a regression version of the focal loss proposed by~\citet{yang2021delving}, where the scaling factor is replaced by a continuous function that maps the absolute error into [0, 1]. Precisely, it can be expressed as:
\begin{equation}
    \mathcal{L}_{Focal-R}=\frac{1}{n}\sum_{i=1}^n\sigma(\left|\beta e_i\right|)^\gamma e_i,
\end{equation}
where $e_i$ denotes the $L_1$ error of the $i$-th sample in training set, $\sigma(\cdot)$ denotes the $Sigmoid$ function, and $\beta,\gamma$ are hyperparemeters.

\textbf{Label Distribution Smoothing (LDS)}~\cite{yang2021delving} convolves a symmetric kernel with the empirical density distribution to extract a kernel-smoothed version that accounts for the overlap in the information of data samples of nearby labels. Given target values $y'$ and any $y$, for $\forall y,y'\in \mathcal{Y}$, LDS indeed computes the density distribution of effective label:
\begin{equation}
    \widetilde{p}(y')\triangleq \int_\mathcal{Y}\text{k}(y,y')p(y)dy,
\end{equation}
where $p(y)$ denotes the appearance number of label $y$ in the training set, $k(\cdot,\cdot)$ denotes the kernel function~(such as the Gaussian kernel function), and $\widetilde{p}(y')$ denotes the effective density of label $y'$.

\textbf{Feature Distribution Smoothing (FDS)}~\cite{yang2021delving} transfers the feature statistics between nearby target bins by performing distribution smoothing on the feature space, thereby calibrating the biased estimates of feature distribution, especially for underrepresented target values. Let $\widetilde{\boldsymbol{\mu}}_b$ and $\widetilde{\boldsymbol{\Sigma}}_b$ be the mean value and covariance of each bin. The smoothed version of the two statistics is as follows:
\begin{gather}
    \widetilde{\boldsymbol{\mu}}_b=\sum_{b'\in\mathcal{B}}\text{k}(y_b,y_{b'})\boldsymbol{\mu}_{b'},\\
    \widetilde{\boldsymbol{\Sigma}}_b=\sum_{b'\in\mathcal{B}}\text{k}(y_b,y_{b'})\boldsymbol{\Sigma}_{b'},
\end{gather}
where $\mathcal{B}$ means the target bins.
% Lastly, Balanced MSE~\cite{BalancedMSE} proposes a novel regression loss to restore a balanced prediction by leveraging the training label distribution prior to make a statistical conversion. 

%% file: Appendix/Appendix-F-Backbone.tex
\section{Backbone models}
How to effectively and efficiently represent molecules is a crucial problem in biology and chemistry. Recently, numerous efforts have since been introduced to obtain better molecular representations~\cite{DBLP:journals/jcisd/RogersH10, weininger1988smiles, wang2019dgl, kipf2017semisupervised}. In general, existing work can be divided into three main categories: conventional molecular fingerprints~(Morgan~\cite{DBLP:journals/jcisd/RogersH10}), string-based representations~(Transformer~\cite{vaswani2017attention}), and graph-based representations~(Graph neural networks~\cite{kipf2017semisupervised}). In what follows, we briefly introduce three main representative methods.
% followed by ablations of backbone models on USPTO-50K. 

\paragraph{Morgan.} Fingerprints is a conventional molecular representation, which applies a kernel to a molecule to generate a numerical vector. Morgan~\cite{DBLP:journals/jcisd/RogersH10} is a representative fingerprint suitable for both small and large molecules by combining substructure and atom‐pair concepts. Morgan is a similarity fingerprint consisting of two atom types: connectivity~(element, \#heavy neighbors, \#Hs, charge, isotope, inRing) and chemical features~(donor, acceptor, aromatic, halogen, basic, acidic). Morgan also takes into account the neighborhood of each atom within less than 3 bonds.

\paragraph{Transformer.} The Transformer architecture has pushed the boundaries of many research domains, such as Neural Language Processing~(NLP) and Computer Vision~(CV). The Transformer layer mainly consists of two components: a self-attention module and a position-wise feed-forward network~(FFN). We denote by $\bm{H} = \left[\bm{h}_1;\cdots;\bm{h}_n\right] \in \mathbb{R}^{N \times d}$ the input matrix of the self-attention module where $d$ is the hidden dimension and $\bm{h}_i$ is the embedding vector at position $i$. We project the input $\bm{H}$ by three learnable weight matrices $\bm{W}_Q \in \mathbb{R}^{d\times d_Q}$, $\bm{W}_K \in \mathbb{R}^{d\times d_K}$, and $\bm{W}_V \in \mathbb{R}^{d\times d_V}$ and obtain the corresponding representations $\bm{Q}$, $\bm{K}$, and $\bm{V}$ respectively. Overall, the self-attention module is calculated as follows:
\begin{equation}
\begin{aligned}
    \bm{Q} = H\bm{W}_{Q}, \quad \bm{K} &= H\bm{W}_{K}, \quad \bm{V} = H\bm{W}_{V}, \\
    \text{Att}(\bm{H}) &= \frac{\bm{Q}\bm{K}^{T}}{\sqrt{d_K}} \bm{V} ,
\end{aligned}
\end{equation}
where the term $\frac{\bm{Q}\bm{K}^{T}}{\sqrt{d_K}}$ usually measures the similarity between queries and keys. We often set $d_Q = d_K = d_V = d$ for simplicity. Typically, we employ multi-head attention layers to stabilize the learning process and enlarge the expressive power of self-attention.

\paragraph{Graph neural networks.} Modern GNNs follows a message-passing mechanism~\cite{wang2019dgl}. During each message-passing iteration, a hidden embedding $\bm{h}_u^{(k)}$ corresponding to each node $u \in \mathcal{V}$ is updated by aggregating information from $u$'s neighborhood $\mathcal{N}(u)$. Graph Convolutional Networks~(GCN)~\cite{kipf2017semisupervised} updates the hidden embedding as
\begin{equation}
    \bm{H}^{(l+1)}=\sigma\left(\hat{\bm{A}} \bm{H}^{(l)} \bm{W}^{(l)}\right),
\end{equation}
where $\bm{H}^{(l+1)} = \left[\bm{h}_1^{(l+1)},\cdots,\bm{h}_n^{(l+1)}\right]$ is the hidden matrix of the $(l+1)$-th layer. 
$\hat{\bm{A}}=\hat{\bm{D}}^{-1 / 2}(\bm{A}+\bm{I}) \hat{\bm{D}}^{-1 / 2}$ is the re-normalization of the adjacency matrix, and $\hat{\bm{D}}$ is the corresponding degree matrix of $\bm{A}+\bm{I}$. 
$\bm{W}^{(l)} \in \mathbb{R}^{C_l \times C_{l-1}}$ is the filter matrix in the $l$-th layer with $C_l$ referring to the size of $l$-th hidden layer and $\sigma(\cdot)$ is a nonlinear function, \textit{e.g.}, ReLU.

% \paragraph{Ablation study.}
% \input{Tables/table-3} 
% In order to compare the three aforementioned backbone models, we conduct experiments on USPTO-50K using the random and standard split. As shown in Table~\ref{table:backbones}, we find that data-driven molecular representation methods~(\textit{i.e.}, Transformer and GCN) significantly outperform the conventional molecular representation methods~(\textit{i.e.}, Morgan). Specifically, the absolute improvement over the GCN baseline model and Morgan baseline model is about 20 percent in terms of \emph{Balanced-Acc}, \emph{Balanced-F1}, and \emph{AUROC}. This suggests that the performance of conventional molecular representation methods will be largely hindered due to the enormous magnitude of possible stable chemical compounds. Moreover, the GCN model can better preserve the rich structural information of molecules and surpass the Transformer models by a large margin.

%% file: Appendix/Appendix-G-Implementation.tex
\section{Implementation Details}

\subsection{Evaluation Metrics}
In this section, we give a description of 23 evaluation metrics we used in ImDrug, including 6 metrics for the regression task, 8 metrics for the binary classification task, 3 multi-class classification metrics, and 6 metrics for the molecule generation task. Besides, we provide 2 novel imbalanced learning metrics, \textit{i.e.}, Balanced Accuracy and Balanced F1. 

\subsubsection{Regression}

\begin{enumerate}[label=\arabic*),leftmargin=*]
\item \textit{Mean Squared Error (MSE)} computes the mean squared error. 
It is defined as 
\begin{equation}
    MSE(y,\hat{y})=\frac{1}{n} \sum_{i=1}^{n} (y_i-\hat{y}_i)^2,
\end{equation}
where $\hat{y}_i$ is the predicted value of $i$-th sample, $y_i$ is the corresponding true value, and $n$ is the number of samples.

\item \textit{Root-Mean Squared Error (RMSE)} computes the root mean squared error.
It is defined as 
\begin{equation}
    RMSE(y,\hat{y})=\sqrt{\frac{1}{n} \sum_{i=1}^{n} (y_i-\hat{y}_i)^2},
\end{equation}
where $\hat{y}_i$ is the predicted value of $i$-th sample, $y_i$ is the corresponding true value, and $n$ is the number of samples.

\item \textit{Mean Absolute Error (MAE)} computes mean absolute error.
It is defined as 
\begin{equation}
    MAE(y,\hat{y})=\frac{1}{n} \sum_{i=1}^{n} |y_i-\hat{y}_i|,
\end{equation} 
where $\hat{y}_i$ is the predicted value of $i$-th sample, $y_i$ is the corresponding true value, and $n$ is the number of samples.

\item \textit{Coefficient of Determination ($\bm{R^2}$)} computes the coefficient of determination, usually denoted as $\bm{R^2}$.
It is defined as 
\begin{equation}
    R^2(y,\hat{y})=1-\frac{\sum_{i=1}^{n}(y_i-\hat{y}_i)^2}{\sum_{i=1}^{n}(y_i-\bar{y})^2},
\end{equation} 
where $\bar{y}=\frac{1}{n}\sum_{i=1}^{n}y_i$.

\item \textit{Pearson Correlation Coefficient (PCC)} computes the amount of linear correlations between the true values and the predicted values.
It is defined as 
\begin{equation}
    PCC=\frac{C_{1,0}}{\sqrt{C_{1,1}*C_{0,0}}},
\end{equation} 
where $C$ is the covariance matrix of two input sequences.

\item \textit{Spearman Correlation Coefficient} computes a Spearman correlation coefficient with associated p-value.
It is defined as 
\begin{equation}
    \rho_{R(y),R(\hat{y})}=\frac{\text{cov}(R(y), R(\hat{y}))}{\sigma_{R(y)}\sigma_{R(\hat{y})}},
\end{equation} 
where $\rho$ denotes the usual Pearson correlation coefficient, but applied to the rank variables, $\text{cov}(R(y), R(\hat{y}))$ is the covariance of the rank variables,  $\sigma_{R(y)}$ and $\sigma_{R(\hat{y})}$ are the standard deviations of the rank variables.
\end{enumerate}

\subsubsection{Binary Classification}

\begin{enumerate}[label=\arabic*),leftmargin=*]
\item \textit{ROC-AUC} computes Area Under the Receiver Operating Characteristic Curve from prediction scores.

\item \textit{PR-AUC/AUPRC} computes the Area Under the Precision-Recall Curve from prediction scores.

\item \textit{Accuracy} computes the accuracy score, either the fraction (default) or the count (normalize=False) of correct predictions.
It is defined as 
\begin{equation}
    Accuracy(y,\hat{y})=\frac{1}{n} \sum_{i=1}^{n} \bm{1}(\hat{y}_i=y_i),
\end{equation} 
where $\hat{y}_i$ is the predicted value of $i$-th sample, $y_i$ is the corresponding true value, and $n$ is the number of samples.

\item \textit{Precision} computes the precision score.
It is defined as 
\begin{equation}
    Precision = \frac{tp}{tp + fp},
\end{equation}
where $tp$ is the number of true positives and $fp$ the number of false positives.

\item \textit{Recall} computes the recall score.
It is defined as 
\begin{equation}
    Recall = \frac{tp}{tp + fn},
\end{equation}
where $tp$ is the number of true positives and $fn$ the number of false negatives.

\item \textit{F1} computes the F1 score, also known as balanced F-score or F-measure.
It is defined as 
\begin{equation}
    F\textit{1} = \frac{2 \times Precision \times Recall}{Precision + Recall}.
\end{equation}

\item \textit{Precision at Recall of K} computes the precision value at the minimum threshold where recall has \textit{K}.

\item \textit{Recall at Precision of K} computes the recall value at the minimum threshold where precision has \textit{K}.

\end{enumerate}

\subsubsection{Multi-calss Classification}

\begin{enumerate}[label=\arabic*),leftmargin=*]
\item \textit{Micro-F1, Micro-Precision, Micro-Recall, Accuracy} computes metrics globally by counting the total true positives, false negatives, and false positives.

\item \textit{Macro-F1} computes metrics for each label and finds their unweighted mean. This does not take label imbalance into account.

\item \textit{Cohen’s Kappa~(Kappa)} is a statistic that measures inter-annotator agreement.
It is defined as
\begin{equation}
    \kappa = \frac{p_o-p_e}{1-p_e},
\end{equation}
where $p_o$ is the empirical probability of agreement on the label assigned to any sample~(the observed agreement ratio), and $p_e$ is the expected agreement when both annotators assign labels randomly. $p_e$ is estimated using a per-annotator empirical prior over the class labels.
\end{enumerate}

\subsubsection{Molecule Generation Metric}

\begin{enumerate}[label=\arabic*),leftmargin=*]
\item \textit{Diversity} evaluates the internal diversity of a set of molecules.

\item \textit{KL divergence} evaluates the KL divergence of the set of generated smiles using the list of training smiles as reference. 

\item \textit{Frechet ChemNet Distance~(FCD)} evaluates the FCD distance between generated smiles set and training smiles set.

\item \textit{Novelty} evaluates the novelty of set of generated smiles using list of training smiles as reference.
It is defined as
\begin{equation}
    Novelty = \frac{|S_{\text{gen}} \setminus S_{\text{train}}|}{|S_{\text{gen}}|},
\end{equation}
where $S_{\text{gen}}$ is the set of generated SMILES strings, $S_{\text{train}}$ is the set of SMILES strings for training, and $\setminus$ means set minus.

\item \textit{Validity} evaluates the chemical validity of a single molecule in terms of SMILES string.
It is defined as
\begin{equation}
    Validity = \frac{|valid(S)|}{|S|},
\end{equation}
where $S$ is the set of SMILES strings, $valid(\cdot)$ removes the SMILES strings failed to be converted as molecules.

\item \textit{Uniqueness} evaluates the uniqueness of a list of SMILES string, i.e., the fraction of unique molecules among a given list.
It is defined as
\begin{equation}
    Uniqueness = \frac{|unique(S)|}{|S|},
\end{equation}
where $S$ is the set of SMILES strings, $unique(\cdot)$ removes the SMILES strings converted to the same molecules as the others.
\end{enumerate}

\subsubsection{Novel Imbalanced Learning Metrics}
\emph{Balanced-Acc} and \emph{Balanced-F1} are two proposed imbalanced learning metrics, which are introduced in Section 3.2
% ~\ref{sec:evaluation_metrics} 
of the main text. We provide a script\footnote{\url{https://github.com/DrugLT/ImDrug/blob/main/rebuttal.ipynb}} in our GitHub repo to demonstrate the three key advantages of our proposed balanced metrics:
\begin{enumerate}
    \item They are the only metrics that are invariant to the label distribution of test sets.
    \item Due to 1, they are the only metrics that can be tested without loss of fairness on much larger, imbalanced test sets, resulting in significantly lower variance/uncertainty.
    \item Due to 2, the lower variance/uncertainty means that when ranking different models, the proposed metrics provide better statistical significance and discriminative power, which is evident in our pairwise t-tests.

\end{enumerate}

\subsection{Encoding Featurizers}
To encourage a diverse development environment, our released benchmark provides 23 featurizers at the bottom level. In what follows, we provide a detailed description of the 23 featurizers for data processing utilities, involving 15 featurizers customized for drugs and 8 featurizers customized for proteins. 

\subsubsection{Drug featurizers.}
\begin{enumerate}[label=\arabic*),leftmargin=*]
\item \textit{Morgan} encodes SMILES strings of drugs into extended-connectivity fingerprints for representing quantitative structure-activity relationship (QSAR).

\item \textit{Pubchem} encodes SMILES strings of drugs into Pubchem substructure-based fingerprints for representing chemical structures.

\item \textit{Daylight} encodes SMILES strings of drugs into Daylight-type fingerprints for representing all possible linkage pathways for drugs to reach a given length.

\item \textit{RDKit\_2d\_normalized} encodes SMILES strings of drugs into normalized descriptastorus by applying a series of normalization transforms to correct functional groups and recombine charges.

\item \textit{ESPF} encodes SMILES strings of drugs into explainable substructure partition fingerprints which can cleverly partition the input drug to discrete pieces of moderate-sized sub-structures.

\item \textit{CNN} encodes SMILES strings of drugs with Convolutional Neural Networks~(CNN) for automatically obtaining low-dimensional representations of input drugs.

\item \textit{CNN\_RNN} encodes SMILES strings of drugs with a Gated Recurrent Unit~(GRU) or Long Short-Term Memory~(LSTM) on top of CNN, which can model the nonlinear order of SMILES strings.

\item \textit{Transformer} encodes SMILES strings of drugs with the transformer on ESPF.

\item \textit{MPNN} encodes SMILES strings of drugs with message passing neural networks, containing the message passing and readout components.

\item \textit{ErG} encodes SMILES strings of drugs with extended reduced graphs for obtaining the pharmacophore-type node descriptions of drugs.

\item \textit{DGL\_GCN} first transforms SMILES strings of drugs to molecular graphs based on the DGL\footnote{https://github.com/dmlc/dgl} library. DGL graphs are then modeled with Graph Convolutional Networks (GCN) to aggregate information from the neighbor nodes.

\item \textit{DGL\_NeuralFP} first transforms SMILES strings of drugs to molecular graphs based on the DGL library and then constructs non-linear fingerprints with neural networks.

\item \textit{DGL\_AttentiveFP} first transforms SMILES strings of drugs to molecular graphs based on the DGL library and then learns interpretable representations of drugs with a graph attentive mechanism.

\item \textit{DGL\_GIN\_AttrMasking} first transforms SMILES strings of drugs to molecular graphs based on the Graph Isomorphism Network~(GIN) and then adapts the pretraining strategy to molecule graph with attribute masking.

\item \textit{DGL\_GIN\_ContextPred} first transforms SMILES strings of drugs to molecular graphs based on the GIN model and then adapts the pretraining strategy to molecule graph with context prediction.
\end{enumerate}

\subsubsection{Protein Featurizer}
\begin{enumerate}[label=\arabic*),leftmargin=*]
\item \textit{AAC} directly encodes an amino acid sequence for representing a target protein.
\item \textit{PseudoAAC} encodes a pseudo-amino acid sequence for representing a target protein.
\item \textit{Conjoint\_triad} encodes the conjoint triad features for representing a protein. Conjoint\_triad first clusters 20 amino acids into seven classes and then regards any three consecutive amino acids into a unit.  
\item \textit{Quasi-seq} represents a target protein by deriving the quasi-sequence order descriptor from the physicochemical distance matrix between the 20 amino acids.
\item \textit{ESPF} encodes an amino acid sequence into an explainable substructure partition, which can cleverly partition the input protein into discrete pieces of moderate-sized sub-structures.
\item \textit{CNN} encodes an amino acid sequence of a target protein with CNN for automatically obtaining low-dimensional representations of input drugs.
\item \textit{CNN\_RNN} encodes an amino acid sequence of a target protein with a GRU or LSTM on top of CNN, which can model the nonlinear order of the amino acid sequence.
\item \textit{Transformer} encodes an amino acid sequence of a target protein with transformer on ESPF.
\end{enumerate}

\subsection{Hyperparameters and Infrastructure}\label{appendix:G.3}
For the reproducibility of our proposed benchmark, we list the hyperparameters used in the ImDrug benchmark. We mainly follow the official hyperparameters based on the DeepPurpose\footnote{https://github.com/kexinhuang12345/DeepPurpose} framework. Specifically, we uniformly sample batches of size $128$ for training with the maximum number of epochs $200$. We adopt an ADAM optimizer with a learning rate of $1e\text{-}3$. The value of momentum and weight decay of ADAM is set to $0.9$ and $2e\text{-}4$ respectively. We use the linear warmup strategy to schedule the learning rate where we linearly increase the learning rate from a low rate to a constant rate thereafter. We set the warmup epoch and the linear factor $0.01$ and $20$ respectively.

Our implementation of ImDrug benchmark is based on the DeepPurpose framework. We implement GNN models based on the DGL\footnote{https://github.com/dmlc/dgl} library. 
We use GCN as the representative GNN model in our experiments. The GCN model consists of three layers, whose hidden feature dimension is 64. The detailed hyperparameters of each backbone model are listed as follows:

\begin{itemize}[leftmargin=14pt]
    \item DGL\_GCN
    \begin{itemize}
        \item Dimension of the hidden layer: 64
        \item Number of layers: 3
        \item Non-linearity function: ReLU
    \end{itemize}
    \item Morgan
    \begin{itemize}
        \item Dimension of the hidden layer: [1024, 256, 64],
    \end{itemize}
    \item Transformer
    \begin{itemize}
        \item Dimension of feature embedding: 128,
        \item Number of attention heads: 8,
        \item Number of layers: 8,
        \item Dropout rate: 0.1,
        \item Dropout rate in the attention layers: 0.1,
        \item Dropout rate in the hidden layers: 0.1,
    \end{itemize}
\end{itemize}

Moreover, all experiments are conducted with the following experimental settings:
\begin{itemize}[leftmargin=14pt]
    \item Operating system: Linux Red Hat 4.8.2-16
    \item CPU: Intel(R) Xeon(R) Platinum 8255C CPU @ 2.50GHz
    \item GPU: NVIDIA Tesla V100 SXM2 32GB
    \item Software versions: Python 3.8.10; Pytorch 1.9.0+cu102; Numpy 1.20.3; SciPy 1.7.1; Pandas 1.3.4; scikit-learn 1.0.1; PyTorch-geometric 2.0.2; DGL 0.7.2; Open Graph Benchmark 1.3.2
\end{itemize}

%% file: Appendix/Appendix-I-Metrics.tex
\section{More Results on Other Evaluation Metrics}
For a comprehensive evaluation, we adopt widely-used metrics for imbalanced datasets including AUPRC, Kappa, MCC, Weighted-F1, Micro-F1, and Macro-F1. As shown in Table~\ref{table:othermertic}, we compare different deep imbalanced learning methods under the random and standard split on four datasets including HIV, SBAP, USPTO-50K, and DrugBank.

We observe that results in Table~\ref{table:othermertic} are relatively consistent with those in Table~\ref{table:overall} and moreover the proposed balanced-Acc and balanced-F1 have two clear advantages in label distribution shift invariance and the significance of evaluation. First, the proposed balanced metrics are invariant to label distribution shift and exhibit a good Pearson's R correlation~(up to 0.9 in average) between random and standard splits on four datasets. AUPRC has the highest Pearson's R correlation among 
the rest metrics, which is 0.7346 in average. Then we explore a key property, p-value, on all metrics to quantify the ability for evaluation significance. Simply put, a smaller p-value represents stronger evidence in favor of the alternative hypothesis. On all four datasets, the proposed balanced metrics achieve consistently lower p-values. On average, p-values of balanced-Acc and balanced-F1 are respectively 2.47\% and 1.01\% lower than those of AUPRC.
\input{Tables-new/table-5-metrics}

%% file: Tables-new/table-5-metrics.tex
% \usepackage{multirow}

\begin{table}[h]
\small
\centering
\caption{\textbf{Results for random and standard splits on 4 ImDrug classification datasets.} We perform binary classification on HIV~(single-instance prediction) and SBAP~(multi-instance prediction), and long-tailed classification on USPTO-50K~(single-instance prediction) and DrugBank~(multi-instance prediction). For each split and metric, the best method is \textbf{bolded} and the second best is \underline{underlined}.}\label{table:othermertic}
\begin{adjustbox}{max width=\textwidth}
\begin{tabular}{ccc|cccccc|cccccc}
\toprule
\multicolumn{3}{c|}{Dataset}  &\multicolumn{6}{c|}{\textbf{HIV}}&\multicolumn{6}{c}{\textbf{SBAP}}
\\ \midrule
\multicolumn{3}{c|}{Metric}                                                                                         & AUPRC                          & Kappa                           & MCC            & Weighted-F1 & Micro-F1 & Macro-F1  & AUPRC                          & Kappa                           & MCC            & Weighted-F1 & Micro-F1 & Macro-F1 \\ \midrule
\multicolumn{1}{c|}{}                            &                                            & Vanilla GCN       & 69.92                        & 44.15                         & 44.48 &  \textbf{96.39}            &     \textbf{96.56}        &  72.06   & \underline{78.17}
& 53.95            & 53.97     & 97.61     & 97.59     & 76.98      \\\cmidrule{2-15}
\multicolumn{1}{c|}{}                            &                                            & BS  & {69.50}                        & {22.12}                        & {27.72} &   90.99          &      87.80       &  59.87   & 77.86    & 28.90    & 39.00     & {92.82}    & {89.80}          &  63.24   \\
\multicolumn{1}{c|}{}                            &                                            & CB\_F  & 69.24                        & 38.78                        & 39.68 &    95.40         &      94.97       &  69.31   & 76.82     & 50.11   & 51.99      & 96.99    & 96.57          &  74.99   \\
\multicolumn{1}{c|}{}                            & \multirow{-3}{*}{\begin{tabular}{c}Class- \\ rebalancing\end{tabular}}        & CS    & {68.78}                        & {24.70}                        & 30.00 &        {91.74}      &      {88.99}       & {61.40}   & 75.87     & 35.89    & 44.46      & 94.45   & 92.43       & 67.26     \\\cmidrule{2-15}
\multicolumn{1}{c|}{}                            &                                            & Mixup              & \textbf{71.25} & \textbf{45.72} & \textbf{45.85} & \underline{96.36} & \underline{96.41} & \textbf{72.85} & \textbf{78.23} & \textbf{55.48} & \textbf{55.52} & \underline{97.69} & \underline{97.68} & \underline{77.74}   \\
\multicolumn{1}{c|}{}                            &                                            & Remix              & \underline{70.96} & {44.59} & \underline{45.09} & {96.09} & {96.09} & {72.30} & {77.71} & \underline{54.54} & \underline{54.52} & \textbf{98.26} & \textbf{98.26} & \textbf{78.48}     \\
\multicolumn{1}{c|}{}                            & \multirow{-3}{*}{\begin{tabular}{c}Information \\augmentation\end{tabular}} & DiVE               & {67.59} & {22.10} & {29.80} & {90.99} & {86.17} & {59.83} & {63.33} & {29.09} & {29.30} & {90.09} & {85.18} & {58.89}  \\\cmidrule{2-15}
\multicolumn{1}{c|}{}                            &                                            & CDT                & {70.49} & \underline{44.65} & {44.71} & {96.27} & {96.29} & \underline{72.33} & {77.64} & {54.28} & {54.31} & {97.62} & {97.59} & {77.14}     \\
\multicolumn{1}{c|}{}                            &                                            & Decoupling         & {69.60} & {37.96} & {38.50} & {95.43} & {95.06} & {68.95} & {76.75} & {52.31} & {53.91} & {97.17} & {96.82} & {76.10}
     \\
\multicolumn{1}{c|}{}                            &                                            & IB & {66.52} & {19.39} & {26.28} & {89.44} & {85.25} & {57.92} & {62.32} & {28.58} & {29.67} & {89.06} & {83.68} & {56.60}     \\
\multicolumn{1}{c|}{\multirow{-11}{*}{Random}}   & \multirow{-4}{*}{\begin{tabular}{c}Module \\Improvement\end{tabular}}       & BBN                & {68.00} & {21.98} & {28.53} & {89.97} & {86.11} & {59.33} & {75.79} & {34.90} & {43.75} & {94.22} & {92.06} & {66.69}    \\ \midrule\midrule
\multicolumn{1}{c|}{}                            &                                            & Vanilla GCN       & {79.30} & {42.49} & {50.17} & {68.99} & {71.24} & {68.99} & {94.49} & {54.58} & {60.15} & {76.21} & {77.29} & {76.21}      \\\cmidrule{2-15}
\multicolumn{1}{c|}{}                            &                                            & BS  & {78.20} & {50.09} & {51.60} & {74.68} & {75.04} & {74.68} & \underline{95.25} & {76.82} & {77.27} & {88.38} & {88.41} & {88.38}    \\
\multicolumn{1}{c|}{}                            &                                            & CB\_F  & {79.37} & {43.14} & {50.05} & {69.57} & {71.57} & {69.57} & {94.79} & {68.17} & {70.90} & {83.78} & {84.08} & {83.78}      \\
\multicolumn{1}{c|}{}                            & \multirow{-3}{*}{\begin{tabular}{c}Class- \\rebalancing\end{tabular}}        & CS    & {79.12} & \textbf{53.89} & \textbf{56.56} & \textbf{76.39} & \textbf{76.95} & \textbf{76.39} & \textbf{95.46} & \textbf{79.57} & \textbf{79.63} & \textbf{89.78} & \textbf{89.79} & \textbf{89.78} \\\cmidrule{2-15}
\multicolumn{1}{c|}{}                            &                                            & Mixup              & \underline{79.65} & {42.96} & {50.71} & {69.31} & {71.48} & {69.31} & {93.67} & {55.08} & {60.69} & {76.48} & {77.54} & {76.48}     \\
\multicolumn{1}{c|}{}                            &                                            & Remix              & {79.28} & {42.05} & {50.77} & {69.99} & {71.84} & {68.23} & {93.38} & {55.96} & {62.39} & {77.98} & {79.09} & {78.95}   \\
\multicolumn{1}{c|}{}                            & \multirow{-3}{*}{\begin{tabular}{c}Information \\augmentation\end{tabular}} & DiVE               &  {79.73} & \underline{51.08} & {52.25} & \underline{76.16} & \underline{76.63} & {75.01} & {91.28} & {70.63} & {71.26} & {86.00} & {86.23} & {86.01} \\\cmidrule{2-15}
\multicolumn{1}{c|}{}                            &                                            & CDT                & {79.12} & {42.48} & {50.52} & {68.94} & {71.24} & {68.94} & {93.69} & {58.38} & {63.21} & {78.40} & {79.19} & {78.40}    \\
\multicolumn{1}{c|}{}                            &                                            & Decoupling         & {78.50} & {46.28} & \underline{52.66} & {71.52} & {73.14} & {71.52} & {93.83} & {62.96} & {66.45} & {80.98} & {81.48} & {80.98}   \\
\multicolumn{1}{c|}{}                            &                                            & IB &  {78.71} & {50.68} & {51.50} & \underline{75.14} & {75.34} & {75.14} & {90.82} & {69.73} & {69.75} & {84.87} & {84.87} & {84.87}   \\
\multicolumn{1}{c|}{\multirow{-11}{*}{Standard}} & \multirow{-4}{*}{\begin{tabular}{c}Module \\Improvement\end{tabular}}       & BBN                & \textbf{80.66} & {46.99} & {47.40} & {73.39} & {73.50} & {73.39} & {94.82} & \underline{78.22} & \underline{78.52} & \underline{89.09} & \underline{89.11} & \underline{89.09}    \\ \bottomrule

\\
\toprule
\multicolumn{3}{c|}{Dataset}  &\multicolumn{6}{c|}{\textbf{USPTO-50K}}&\multicolumn{6}{c}{\textbf{DrugBank}}
\\ \midrule
\multicolumn{3}{c|}{Metric}                                                                                         & AUPRC                          & Kappa                           & MCC            & Weighted-F1 & Micro-F1 & Macro-F1  & AUPRC                          & Kappa                           & MCC            & Weighted-F1 & Micro-F1 & Macro-F1 \\ \midrule
\multicolumn{1}{c|}{}                            &                                            & Vanilla GCN       & {96.55} & {95.00} & {95.01} & {95.99} & {96.00} & {92.70} & {93.91} & {93.04} & {93.04} & {94.12} & {94.13} & {88.47}      \\\cmidrule{2-15}
\multicolumn{1}{c|}{}                            &                                            & BS  & {96.61} & {94.75} & {94.75} & {95.83} & {95.79} & {92.24} & {94.85} & {91.10} & {91.14} & {92.86} & {92.42} & {72.22}  \\
\multicolumn{1}{c|}{}                            &                                            & CB\_F  & {95.23} & {94.28} & {94.29} & {95.47} & {95.42} & {92.26} & {70.95} & {82.18} & {82.22} & {85.13} & {84.92} & {80.63}  \\
\multicolumn{1}{c|}{}                            & \multirow{-3}{*}{\begin{tabular}{c}Class- \\ rebalancing\end{tabular}}        & CS    & {96.42} & {93.73} & {93.74} & {95.01} & {94.97} & {92.25} & {89.24} & {72.10} & {72.57} & {75.83} & {75.57} & {71.19}      \\\cmidrule{2-15}
\multicolumn{1}{c|}{}                            &                                            & Mixup              & \textbf{97.27} & \underline{96.34} & \underline{96.34} & \underline{97.09} & \underline{97.07} & \underline{94.44} & \underline{96.24} & \underline{95.37} & \underline{95.37} & \underline{96.09} & \underline{96.10} & \underline{92.87}   \\
\multicolumn{1}{c|}{}                            &                                            & Remix              & \underline{97.22} & \textbf{96.47} & \textbf{96.48} & \textbf{97.21} & \textbf{97.18} & \textbf{94.55} & \textbf{97.33} & \textbf{95.70} & \textbf{95.71} & \textbf{96.37} & \textbf{96.38} & \textbf{95.17}    \\
\multicolumn{1}{c|}{}                            & \multirow{-3}{*}{\begin{tabular}{c}Information \\augmentation\end{tabular}} & DiVE               & {96.21} & {94.68} & {94.68} & {95.79} & {95.70} & {92.79} & {90.36} & {80.22} & {80.30} & {83.45} & {83.29} & {79.40}  \\\cmidrule{2-15}
\multicolumn{1}{c|}{}                            &                                            & CDT                & {96.78} & {94.95} & {94.95} & {95.98} & {95.95} & {92.70} & {82.48} & {70.54} & {71.63} & {78.77} & {73.50} & {51.45}       \\
\multicolumn{1}{c|}{}                            &                                            & Decoupling         & {95.34} & {93.14} & {93.14} & {94.50} & {94.50} & {90.98} & {93.44} & {90.81} & {90.81} & {92.25} & {92.24} & {88.40} 
     \\
\multicolumn{1}{c|}{}                            &                                            & IB & {94.38} & {92.44} & {92.47} & {92.45} & {92.25} & {87.56} & {82.68} & {73.86} & {74.33} & {79.04} & {77.06} & {55.53}     \\
\multicolumn{1}{c|}{\multirow{-11}{*}{Random}}   & \multirow{-4}{*}{\begin{tabular}{c}Module \\Improvement\end{tabular}}       & BBN                & {96.72} & {94.11} & {94.13} & {95.40} & {95.26} & {92.14} & {92.24} & {90.32} & {90.33} & {92.38} & {91.81} & {83.07}   \\ \midrule\midrule
\multicolumn{1}{c|}{}                            &                                            & Vanilla GCN       & {97.31} & {91.28} & {91.35} & {92.15} & {92.15} & {92.15} & {95.15} & {89.25} & {89.41} & {86.96} & {89.38} & {86.96}       \\\cmidrule{2-15}
\multicolumn{1}{c|}{}                            &                                            & BS  & {97.77} & {92.91} & {92.93} & {93.59} & {93.62} & {93.59} & \underline{96.47} & \underline{94.63} & \underline{94.69} & {93.11} & \underline{94.69} & {92.58}     \\
\multicolumn{1}{c|}{}                            &                                            & CB\_F  & {96.78} & {93.41} & {93.44} & {94.11} & {94.07} & {94.11} & {80.75} & {90.44} & {90.59} & {88.60} & {90.55} & {88.09}       \\
\multicolumn{1}{c|}{}                            & \multirow{-3}{*}{\begin{tabular}{c}Class- \\rebalancing\end{tabular}}        & CS    & {97.79} & {91.93} & {91.97} & {92.76} & {92.73} & {92.76} & {95.74} & \underline{94.63} & \underline{94.69} & \underline{93.42} & \underline{94.69} & \underline{92.89} \\\cmidrule{2-15}
\multicolumn{1}{c|}{}                            &                                            & Mixup              & \underline{98.55} & {94.68} & {94.72} & {95.19} & {95.21} & {95.19} & \textbf{97.15} & {85.71} & {86.15} & {83.48} & {85.88} & {83.48}      \\
\multicolumn{1}{c|}{}                            &                                            & Remix              & {98.45} & \underline{94.74} & \underline{94.77} & \underline{95.25} & \underline{95.27} & \underline{95.25} & {96.38} & {89.81} & {90.00} & {87.46} & {89.93} & {87.46}   \\
\multicolumn{1}{c|}{}                            & \multirow{-3}{*}{\begin{tabular}{c}Information \\augmentation\end{tabular}} & DiVE               &  {97.56} & {93.36} & {93.40} & {94.00} & {93.97} & {94.00} & {93.72} & {89.83} & {90.00} & {87.97} & {89.95} & {87.46}  \\\cmidrule{2-15}
\multicolumn{1}{c|}{}                            &                                            & CDT                & {97.88} & {93.10} & {93.13} & {93.78} & {93.79} & {93.78} & {93.96} & {91.04} & {91.12} & {88.78} & {91.15} & {88.29}    \\
\multicolumn{1}{c|}{}                            &                                            & Decoupling         & {95.45} & {87.67} & {87.81} & {87.11} & {87.20} & {87.11} & {94.72} & {89.85} & {90.00} & {87.78} & {89.97} & {87.30}  \\
\multicolumn{1}{c|}{}                            &                                            & IB & {91.74} & {86.86} & {86.94} & {88.08} & {88.17} & {88.08} & {95.13} & {89.86} & {89.96} & {87.61} & {89.98} & {87.14}   \\
\multicolumn{1}{c|}{\multirow{-11}{*}{Standard}} & \multirow{-4}{*}{\begin{tabular}{c}Module \\Improvement\end{tabular}}       & BBN                & \textbf{98.76} & \textbf{95.82} & \textbf{95.82} & \textbf{96.25} & \textbf{96.23} & \textbf{96.25} & {95.69} & \textbf{95.23} & \textbf{95.30} & \textbf{94.14} & \textbf{95.28} & \textbf{93.61}     \\ \bottomrule
\end{tabular}
\end{adjustbox}
\end{table}

%% file: Appendix/Appendix-J-Train-curve.tex
\section{Learning Curves}
\begin{figure}[h]
    \centering
    \includegraphics[width=\columnwidth]{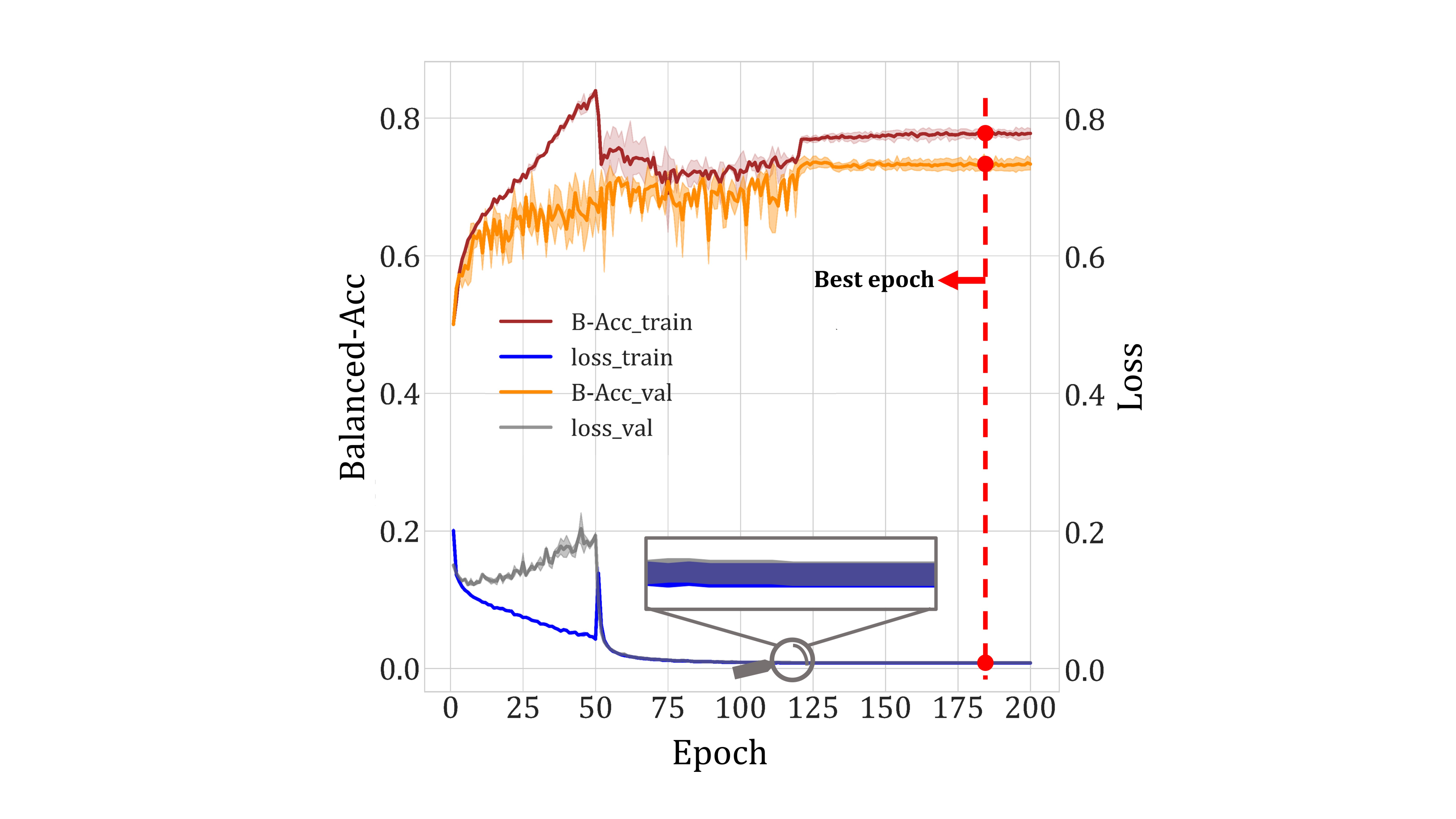}
    \caption{The training/validation curves on HIV dataset with GCN as backbone and \textbf{IB} as baseline. The curves are depicted on three runs.}
    \label{fig:train-curve}
\end{figure}

We show an example of learning curves in Figure~\ref{fig:train-curve} to verify the absence of overfitting issues. The following observations for training on HIV dataset are made:~(1)~The train/validation curves are stable, especially, after 50 epochs, the curves of train and validation loss are highly matched, indicating the basic ability for GCN to stabilize the imbalanced learning; (2)~The balanced-Acc of validation data shows a upward trend as the epoch increases and at the best epoch, the final validation balanced-Acc is only 3.26\% lower than the training balanced-Acc, suggesting no risk of overfitting.
